\documentclass{article}
\usepackage{iclr2027_conference,times}

\usepackage{amsmath,amsfonts,bm}

\def\eqref#1{equation~\ref{#1}}

\def\1{\bm{1}}

\DeclareMathAlphabet{\mathsfit}{\encodingdefault}{\sfdefault}{m}{sl}
\SetMathAlphabet{\mathsfit}{bold}{\encodingdefault}{\sfdefault}{bx}{n}

\usepackage[T1]{fontenc}
\usepackage{microtype}
\usepackage{hyperref}
\usepackage{url}
\usepackage{xurl}
\usepackage{booktabs}
\usepackage{array}
\usepackage{tabularx}
\usepackage{multirow}
\usepackage{graphicx}
\usepackage{wrapfig}
\usepackage{amsmath}
\usepackage{amssymb}
\usepackage{placeins}
\usepackage{listings}
\usepackage{xcolor}
\usepackage{pifont}
\usepackage[most]{tcolorbox}

\definecolor{tplhead}{HTML}{595959}
\definecolor{tplbodybg}{HTML}{F2F2F2}
\definecolor{tpltokctrl}{HTML}{123f73}
\definecolor{tpltokrole}{HTML}{c0392b}
\definecolor{tpltokthink}{HTML}{2980b9}
\definecolor{tpltokesc}{HTML}{a3a3a3}
\newtcolorbox{tplbox}[1]{%
  enhanced, colback=white, colframe=tplhead, colbacktitle=tplhead,
  coltitle=white, fonttitle=\bfseries\footnotesize, title={#1},
  boxrule=0.9pt, titlerule=0pt, arc=2pt,
  left=5pt, right=5pt, top=3pt, bottom=3pt,
  equal height group=tplex,
}
\newtcolorbox{tplinput}{%
  colback=tplbodybg, colframe=tplbodybg, boxrule=0pt, arc=1.5pt,
  left=4pt, right=4pt, top=2pt, bottom=2pt,
  before skip=3pt, after skip=3pt,
}

\newtcolorbox{rollbox}[1]{%
  enhanced, colback=white, colframe=tplhead, colbacktitle=tplhead,
  coltitle=white, fonttitle=\bfseries\footnotesize, title={#1},
  boxrule=0.9pt, titlerule=0pt, arc=2pt,
  left=5pt, right=5pt, top=3pt, bottom=3pt,
}
\newcommand{\rollq}[1]{{\bfseries #1}}
\newcommand{\rolla}[1]{{\color{tpltokctrl}\bfseries #1}}

\newcommand{\rollmuted}[1]{{\color{tpltokesc}#1}}

\lstdefinestyle{prompt}{
  basicstyle=\ttfamily\small,
  breaklines=true,
  frame=single,
  rulecolor=\color{gray!40},
  backgroundcolor=\color{gray!5},
  xleftmargin=4pt,
  xrightmargin=4pt,
  aboveskip=6pt,
  belowskip=6pt,
  columns=fullflexible,
  keepspaces=true,
}

\title{Data-Free On-Policy Distillation:\\
How Far Can We Go Without External Data?}

\author{Gengsheng Li$^{1,2,3,*}$, Mao Zheng$^{3,*}$, Mingyang Song$^{3,*}$, Jie Sun$^{3}$, Zeyuan Liu$^{3}$, \\
\textbf{Ruiqi Liu$^{2}$, Tianyu Yang$^{1,2}$, Qiyong Zhong$^{3}$, Haiyun Guo$^{1,2,\dagger}$, Junfeng Fang$^{4,\dagger}$,} \\
\textbf{Shiming Xiang$^{2}$, Jinqiao Wang$^{1,2,5}$, Tat-Seng Chua$^{4}$} \\
$^{1}$Foundation Model Research Center, Institute of Automation, Chinese Academy of Sciences \\
$^{2}$School of Artificial Intelligence, University of Chinese Academy of Sciences \\
$^{3}$Foundation Model Department, Tencent \\
$^{4}$National University of Singapore \quad $^{5}$Wuhan AI Research \\
$^{*}$Equal contribution \quad $^{\dagger}$Corresponding authors
}

\hypersetup{pdftitle={Data-Free On-Policy Distillation: How Far Can We Go Without External Data?},pdfauthor={}}

\iclrfinalcopy
\begin{document}
\maketitle
\lhead{Preprint}
\suppressfloats[t]
\begin{abstract}
On-policy distillation (OPD) is increasingly applied to frontier foundation model post-training. Prior work in this area has largely focused on algorithmic advances, yet it remains unclear how much OPD depends on its training questions and, in particular, how far this dependence can be reduced. Across two representative single-teacher OPD settings, we find that training on 8 real prompts yields performance comparable to training on 17k problems, while datasets differing substantially in measured difficulty and initial distillation gap yield similar outcomes. Our analyses suggest two complementary explanations: repeated sampling could allow even a few prompts to expose substantial teacher supervision, 
while OPD transfers generalizable reasoning capabilities beyond dataset-specific knowledge. 
Building on these observations, we next propose a data-free on-policy distillation (DF-OPD) setting to investigate whether the system can supply the training questions itself, eliminating the need for external data. With 64 self-generated questions obtained without seed examples, DF-OPD yields performance comparable to full-data OPD in both single-teacher settings. This finding also holds in multi-teacher OPD: across mathematics, code, and instruction following, 1k generated questions achieve performance comparable to training on approximately 7k real post-training examples. We further explore whether OPD can operate even without explicit training questions. The experiments show that this is effective only in limited cases, where the student unexpectedly generates and answers its own questions, thereby reducing the process to an implicit form of DF-OPD. Together, these findings invite a reassessment of the role of training data in on-policy distillation.
Code is available at \url{https://github.com/Ryuki661/DF-OPD}.
\end{abstract}

\section{Introduction}
\label{sec:intro}

\begin{figure}[t]
\centering
\includegraphics[width=\linewidth]{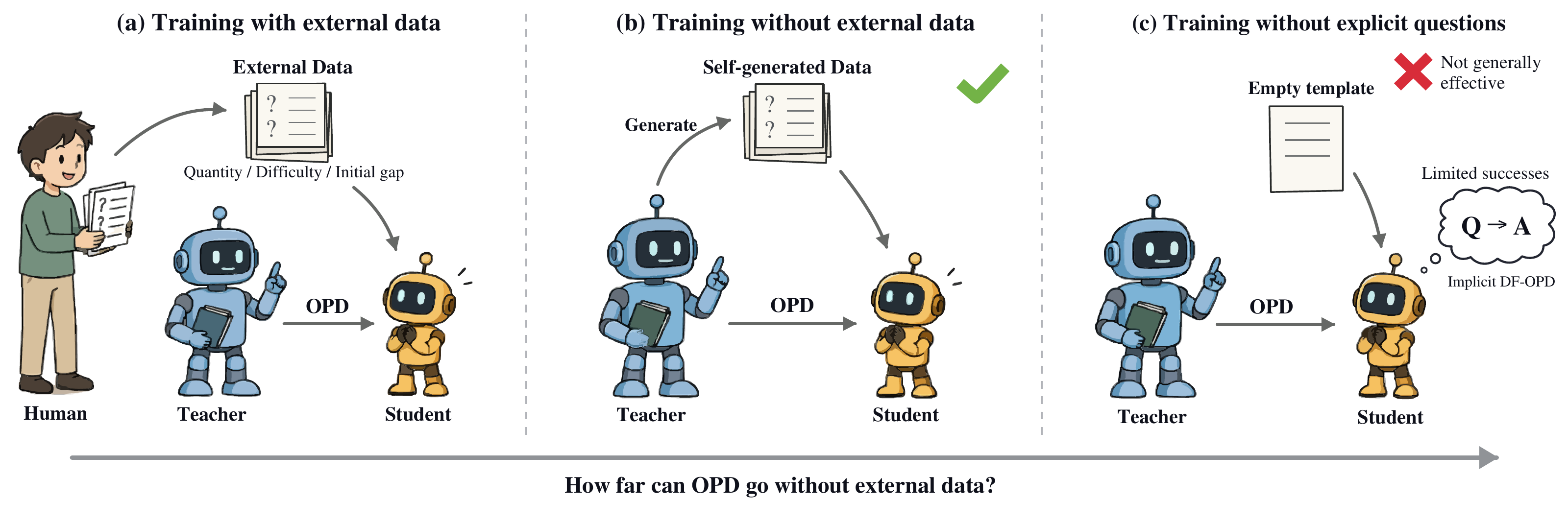}
\caption{\textbf{Progressively reducing OPD's dependence on training questions.} (a)~With external questions, OPD shows limited sensitivity to their quantity, difficulty, and initial distillation gap. (b)~DF-OPD replaces them with self-generated questions, produced by the teacher without seed examples, and performs comparably to full-data OPD. (c)~Empty prompts are effective only in limited cases, where the student spontaneously poses and answers its own question (Self-QA), reducing the process to implicit DF-OPD.}
\label{fig:overview}
\end{figure}

On-policy distillation (OPD) has become a widely adopted component of post-training pipelines for frontier foundation 
models~\citep{qwen3,mimov2flash,deepseekv4}. 
Supervised fine-tuning (SFT) provides dense token-level supervision but suffers from exposure bias, as the model is trained on reference prefixes rather than its own generations. Reinforcement learning with verifiable rewards (RLVR) mitigates this mismatch by learning from model-generated trajectories, but typically relies on sparse outcome-level rewards. OPD combines the advantages of both by providing token-level teacher feedback on student-generated trajectories, bringing together on-policy learning and dense supervision~\citep{minillm,gkd,thinkingmachineslab}.

Prior work has largely focused on algorithmic improvements, including supervision design and optimization~\citep{eopd,exopd,opdeltad,trustopd,trustpd}. Yet it remains relatively unclear how much OPD depends on its training data and, in particular, how far this dependence can be reduced. We investigate this dependence by first examining how sensitive OPD is to its training questions, then asking whether self-generated questions can support effective OPD without external training data, and finally whether explicit questions are needed at all (Figure~\ref{fig:overview}).

\textbf{How sensitive is OPD to its training questions?}
We first study two representative single-teacher OPD settings: one with a teacher obtained by RL-tuning the student, and the other with a larger teacher from the same model family. In both settings, 8 questions yield performance comparable to training on approximately 17k problems. Datasets with substantially different measured difficulty and initial distillation gap also produce similar outcomes. Two analyses help interpret these observations. First, repeated sampling allows a few questions to expose substantial teacher supervision, while the additional signal coverage obtained from more questions exhibits diminishing returns. Second, training on competitive-programming questions recovers over 90\% of the in-domain gains on mathematics benchmarks, 
which suggests that OPD can transfer generalizable reasoning capabilities beyond dataset-specific knowledge.

\textbf{Can self-generated questions replace external data?}
To answer this question, we propose \emph{Data-free On-Policy Distillation} (DF-OPD), a setting in which the system generates its own training questions, eliminating the need for external question corpora during distillation. Specifically, the teacher generates training questions directly from simple instructions, without external seed examples. With 64 self-generated questions, 
DF-OPD performs comparably to full-data OPD trained on approximately 17k real questions, with peak macro-average accuracies of 56.61 versus 56.41 in the RL-tuned teacher setting and 49.16 versus 47.36 in the larger teacher setting.
This finding also holds in multi-teacher OPD (MOPD), where each domain teacher generates questions for its own domain. In this setting, 1k generated questions achieve an overall score of 31.31 across mathematics, code, and instruction following, compared with 31.18 obtained using approximately 7k real post-training examples.

\textbf{Can OPD work without explicit training questions?}
We further remove explicit question content from the input supplied to the student, leaving an empty user prompt without task information~\citep{oneshotopd}. 
Experiments show that this is effective only in limited settings, where the student exhibits unexpected Self-QA behavior: it spontaneously poses a question and attempts to answer it within the same rollout. 
This amounts to an implicit form of DF-OPD with question generation hidden within student rollouts. Further analysis shows that the effectiveness of empty-prompt OPD is associated with the student's Self-QA frequency.
In the RL-tuned teacher setting, the Self-QA rate is 62.0\%, and empty-prompt OPD retains approximately 95.6\% of the mathematics accuracy gain achieved by full-data OPD. 
In contrast, in the larger teacher setting, the Self-QA rate is only 0.1\%, and empty-prompt OPD retains just 7.4\% of the corresponding gain, yielding little improvement over the initial student. This reliance on student-specific behavior limits its generality.

Together, our findings reveal OPD's limited sensitivity to training-data quantity, difficulty, and initial distillation gap, and clarify how far it can operate independently of external training data. 
In particular, DF-OPD points to an intriguing possibility that the system can supply its own training questions and supervision while achieving performance comparable to full-data OPD. These results invite a reassessment of the role of training data in on-policy distillation.

\section{Preliminaries and Experimental Setup}
\label{sec:background}
\subsection{On-Policy Distillation}
\label{sec:opd}
Given a frozen teacher $\pi_\phi$ and a trainable student $\pi_\theta$, OPD applies teacher supervision to trajectories sampled by the student~\citep{minillm,gkd}.
For a training question $x\sim\mathcal D$ and response $y\sim\pi_\theta(\cdot\mid x)$, let $s_t=(x,y_{<t})$ denote the autoregressive state. The reverse-KL objective is
\begin{equation}
\label{eq:opd}
\mathcal L_{\mathrm{OPD}}(\theta)=
\mathbb E_{x\sim\mathcal D,\,y\sim\pi_\theta(\cdot\mid x)}
\left[\sum_{t=1}^{|y|}\mathrm{KL}\!\left(\pi_\theta(\cdot\mid s_t)\,\|\,\pi_\phi(\cdot\mid s_t)\right)\right].
\end{equation}
A question therefore supplies supervision through the states visited along its rollouts, rather than through a fixed reference answer.
In this work, we use a top-$k$ surrogate to balance computational efficiency and supervision density~\citep{rethinkingopd,revisitingopd}.
MOPD, which distills knowledge from multiple domain-specific teachers into a single student, associates each question with a domain label $d(x)$ and replaces $\pi_\phi$ by the corresponding teacher $\pi_{\phi_{d(x)}}$.
See Appendix~\ref{sec:app-opd} for more details.

\subsection{Experimental Setup}
\label{sec:setup}

\textbf{Models.}
We study two representative single-teacher settings:
\emph{RL-tuned teacher}, with DeepSeek-R1-Distill-Qwen-1.5B as the student and its RL-tuned counterpart JustRL-DeepSeek-1.5B as the teacher~\citep{deepseekr1,justrl};
and \emph{larger teacher}, with Qwen3-1.7B as the student and Qwen3-30B-A3B-Instruct-2507 as the teacher~\citep{qwen3}.
For MOPD, we use the released Open-MOPD checkpoints~\citep{openmopd}.
The student is SmolLM3-3B-MixSFT, and the teachers are its three domain-specific RL-tuned variants (SmolLM3-3B-RL-Math/Code/IF) for mathematics, code, and instruction following, respectively.
See Appendix~\ref{sec:app-models} for more model details.

\textbf{Datasets.}
For single-teacher experiments, DAPO-Math-17k (DAPO)~\citep{dapo} serves as the primary source of external training questions.
We additionally use a subset of DeepMath~\citep{deepmath} and Open-Reasoner-Zero (ORZ)~\citep{orz} to examine sensitivity to the data source, and a subset of DeepCoder~\citep{deepcoder} with mathematics-related problems removed to study transfer from programming to mathematics.
For MOPD, the external-data baseline uses a subset of the RL training prompts released by Open-MOPD.
Appendix~\ref{sec:app-datasets} details the data sources and preparation.
Self-generated questions are introduced in Section~\ref{sec:selfgen}.

\textbf{Training and Evaluation.}
All experiments use verl~\citep{verl} and the AdamW~\citep{adamw} optimizer with a learning rate of $10^{-6}$ and a weight decay of $0.01$.
Each training batch contains 64 questions. For question pools smaller than a batch, we repeat questions to fill each batch.
In the single-teacher settings, we evaluate on six mathematics benchmarks:
AIME24~\citep{aime2024}, AIME25~\citep{aime2025}, AMC23~\citep{amc2023}, MATH-500~\citep{mathdataset,math500}, Minerva~\citep{minerva}, and Olympiad-Bench~\citep{olympiadbench}, using $\mathrm{avg}@4$.
For MOPD, we evaluate mathematics on AIME25 and AIME26~\citep{aime26} ($\mathrm{avg}@16$), code on LiveCodeBench v5 and v6~\citep{livecodebench} ($\mathrm{avg}@5$), and instruction following on IFEval~\citep{ifeval} and IFBench~\citep{ifbench}; each domain score averages its two benchmarks.
In each setting, the main metric is the macro average of all benchmark scores.
We report its peak across evaluated checkpoints.
Appendix~\ref{sec:app-training} provides the full training and evaluation configurations.

\section{How Sensitive Is OPD to Its Training Data?}
\label{sec:sensitivity}

OPD obtains teacher supervision along student-generated trajectories, with training questions serving as their starting points. How much does effective distillation depend on the number and properties of these questions? We investigate this across two single-teacher settings, varying question-pool size and comparing data sources that differ in measured difficulty and initial distillation gap.

\subsection{Sensitivity to the Number of Training Questions}
\label{sec:quantity}

\begin{figure}[t]
\centering
\includegraphics[width=\linewidth]{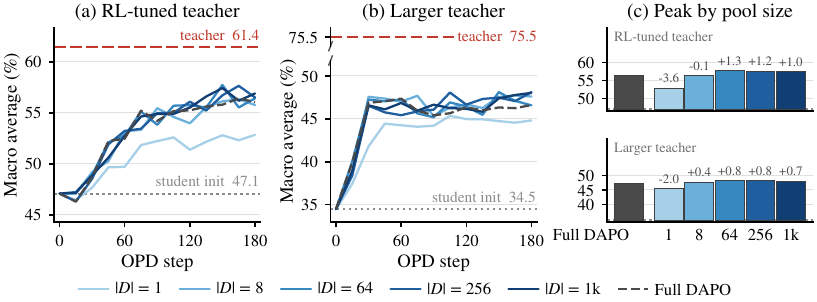}
\caption{\textbf{Few real questions support effective OPD.} (a,b) DAPO training curves under matched update and trajectory budgets. Macro accuracy (\%) is averaged over six mathematics benchmarks, evaluated every 15 updates. (c) Peak scores; bars start at the initial student, and numbers give the difference from full DAPO in percentage points.}
\label{fig:quantity}
\end{figure}

We randomly sample DAPO subsets of size $|D|\in\{1,8,64,256,1\mathrm{k}\}$ and compare them with the 17k full-data DAPO in both single-teacher settings, using the same training and evaluation protocol.

As shown in Figure~\ref{fig:quantity}, just 8 questions already yield performance comparable to the 17k full-data DAPO baseline: 56.3 versus 56.4 in the RL-tuned teacher setting and 47.7 versus 47.4 in the larger teacher setting. This represents a more than 2,000-fold reduction in the number of training questions while maintaining comparable performance. We further find that even with a single question, OPD retains approximately 62\% and 84\% of the full-data gains over the initial students in the RL-tuned teacher and larger teacher settings, respectively. These results show that even very few training questions can suffice for effective OPD.

\subsection{Sensitivity to Data Difficulty and Initial Distillation Gap}
\label{sec:difficulty}

\begin{figure}[t]
\centering
\includegraphics[width=\linewidth]{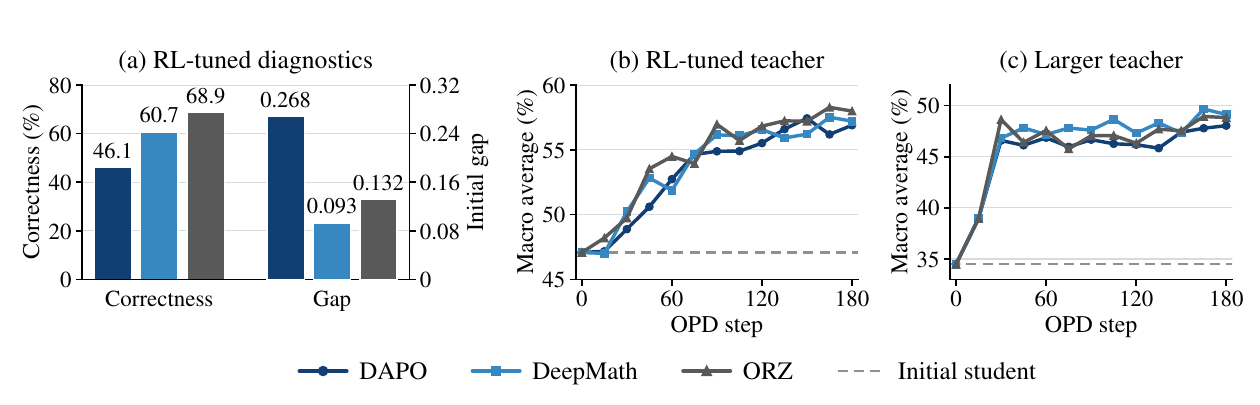}
\caption{\textbf{Initial diagnostics and learning curves.} (a) Initial correctness (left axis) and initial distillation gap (right axis) in the RL-tuned teacher setting. (b,c) Six-benchmark macro accuracy for the RL-tuned and larger teacher settings with 1,000 questions per source.}
\label{fig:difficulty}
\end{figure}

Beyond data quantity, another question worth examining is how sensitive OPD is to data difficulty and initial distillation gap. Here, difficulty is assessed using initial student correctness, while the initial distillation gap measures the KL-based discrepancy (Appendix~\ref{sec:app-gap}) between teacher and student predictions along student-generated trajectories before training. Both metrics are evaluated in the RL-tuned teacher setting. To examine this sensitivity, we compare DAPO, DeepMath, and ORZ in both single-teacher settings, using the subsets of size $|D|=1\mathrm{k}$ randomly sampled from each source.

As shown in Figure~\ref{fig:difficulty}(a), initial correctness is 46.1\%, 60.7\%, and 68.9\% for DAPO, DeepMath, and ORZ, respectively, while the corresponding initial distillation gaps are 0.268, 0.093, and 0.132. Despite these differences, including a nearly threefold difference in the initial distillation gap, the three sources yield similar peak performance in both settings: 57.4--58.3\% in the RL-tuned teacher setting and 48.0--49.7\% in the larger teacher setting (Figure~\ref{fig:difficulty}(b,c)). These results suggest that OPD is relatively insensitive to data difficulty and initial distillation gap.

\subsection{Analysis}
\label{sec:why}

We analyze OPD's limited sensitivity to data quantity, difficulty, and initial distillation gap from two complementary perspectives.

\begin{figure}[t]
\centering
\includegraphics[width=\linewidth]{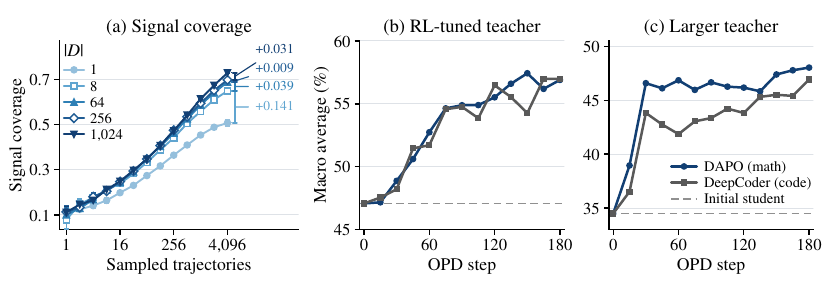}
\caption{\textbf{Signal coverage and cross-domain transfer.} (a) Signal coverage grows with repeated sampling. Mean $\pm$ std over six resampling draws of the frozen initial student; brackets mark gains between adjacent pool sizes at 4,096 trajectories. (b,c) Transfer from programming to mathematics for the RL-tuned and larger teacher settings with $|D|=1{,}000$ questions. Macro accuracy on the same six mathematics benchmarks; dashed lines mark initial students.}
\label{fig:coverage-crossdomain}
\end{figure}

\paragraph{A Few Questions Can Expose Substantial Teacher Supervision.}
\label{sec:coverage}
OPD receives teacher feedback at the states visited along student-generated trajectories. Repeated on-policy sampling from the same question can expose different prefixes and thereby yield different supervision signals. We quantify this using \emph{signal coverage}, which measures how well supervision signals from sampled trajectories represent those in a held-out reference set. Specifically, we represent the supervision signal at each state by a normalized vector combining the student's hidden representation with a correction signal derived from a KL-based discrepancy between the teacher and student. We then compute coverage by measuring how well the reference signal vectors can be reconstructed as linear combinations of the sampled signal vectors. The score ranges from 0 to 1: higher values indicate better representation of the reference supervision, while lower values indicate that more of it remains unrepresented. See Appendix~\ref{sec:app-coverage-definition} for the detailed definition.

For a single question, coverage increases from 0.114 with 1 trajectory to 0.507 with 4,096 trajectories (Figure~\ref{fig:coverage-crossdomain}(a)). In the latter budget, expanding the question pool from 1 to 8 adds 0.141 coverage, whereas expanding it further from 8 to 1,024 adds only about 0.08. These observations suggest that repeated on-policy sampling can expose substantial teacher supervision from very few questions, while enlarging the question pool yields diminishing coverage gains. This offers an explanation for why a small question pool can support effective OPD.

\paragraph{OPD Can Transfer Reasoning Capabilities Beyond Dataset-Specific Knowledge.}
\label{sec:reasoning}
To help explain OPD's limited sensitivity to training data sources, we take the comparison a step further by replacing DAPO mathematics questions with math-filtered coding problems from DeepCoder. These problems draw on different domain knowledge but also require reasoning. Both conditions use $|D|=1\mathrm{k}$ questions for training, while evaluation remains on the same six mathematics benchmarks.

Despite the change in training domain, OPD retains 96\% and 92\% of the gains achieved with DAPO over the initial students in the RL-tuned teacher and larger teacher settings, respectively (Figure~\ref{fig:coverage-crossdomain}(b,c)). The corresponding peak mathematics accuracies are only 0.44 and 1.11 percentage points below those obtained with DAPO. These results suggest that OPD can transfer reasoning capabilities from the teacher beyond dataset-specific knowledge, helping explain its limited sensitivity to training data sources.

\section{Can Self-Generated Questions Replace External Data?}
\label{sec:dfopd}

The preceding experiments show that few questions can support effective OPD and that OPD is relatively insensitive to the data source. This raises a further question: can the distillation system generate its own training questions, eliminating the need for external data? We introduce DF-OPD and examine its effectiveness in both single-teacher and multi-teacher settings.

\subsection{Data-free On-Policy Distillation}
\label{sec:selfgen}
\label{sec:single-results}

\begin{figure}[htbp]
\centering
\includegraphics[width=\linewidth]{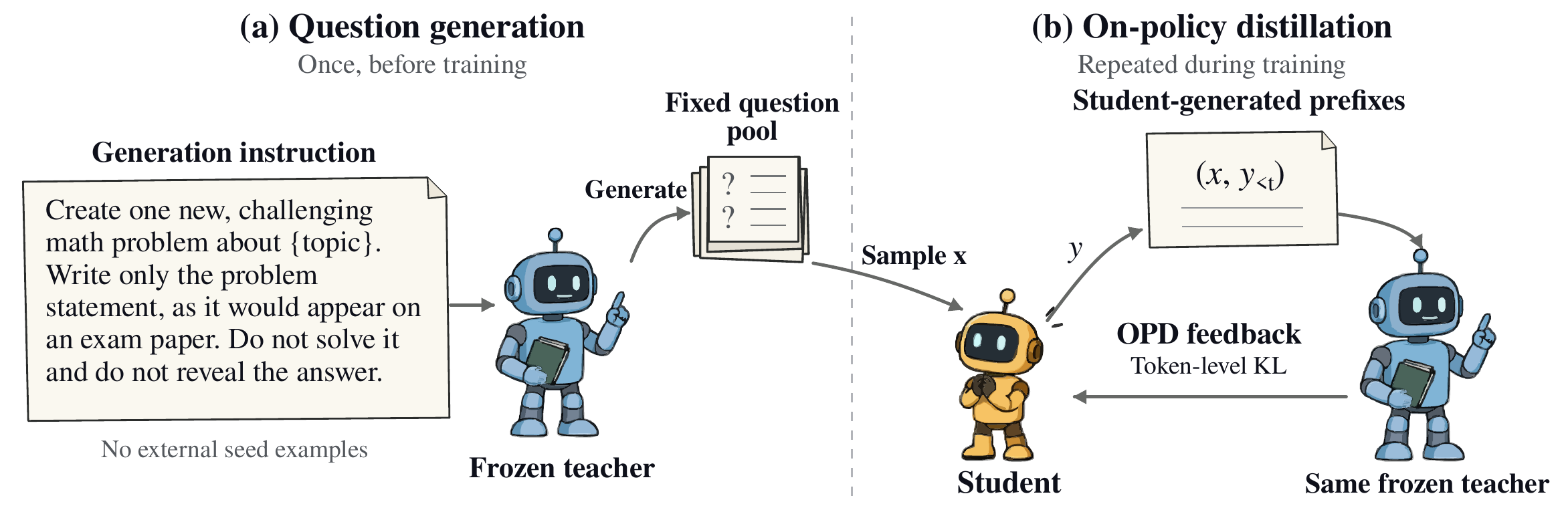}
\caption{\textbf{DF-OPD workflow.} A frozen teacher generates a fixed question pool from a simple instruction. The student samples responses, and the same teacher provides token-level feedback on student-generated prefixes. Standard OPD updates only the student. The instruction shown is the RL-tuned teacher's generation prompt.}
\label{fig:dfopd-workflow}
\end{figure}

In DF-OPD, the teacher supplies both training questions and distillation supervision. Specifically, before distillation begins, we use simple instructions specifying the domain and topic to prompt the teacher to generate a fixed question pool, without external seed examples. The student then samples responses to these questions, and the frozen teacher provides token-level feedback on student-generated prefixes to update the student parameters (Figure~\ref{fig:dfopd-workflow}). The standard OPD objective remains unchanged; only the training questions are replaced with those generated by the teacher itself.

We apply only question extraction, format checks, and deduplication, without verifying correctness or filtering questions based on difficulty or solution-quality scores. Full generation instructions and processing details appear in Appendix~\ref{sec:app-selfgen}. Here, \emph{data-free} means that the distillation system supplies its own training questions and supervision, allowing it to operate without external data.

\begin{figure}[t]
\centering
\includegraphics[width=\linewidth]{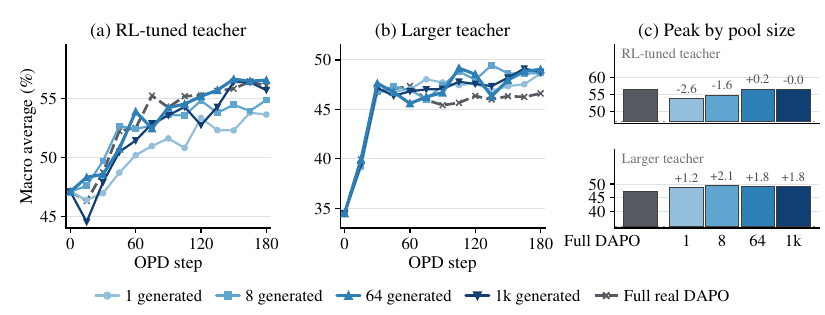}
\caption{\textbf{Self-generated questions replace the external corpus.} (a,b) Single-teacher OPD with different numbers of generated questions, compared with training on the full real DAPO corpus. (c) Peak scores; bars start at the initial student, and numbers give the difference from full real DAPO in percentage points. In both settings, 64 questions reach comparable performance. Complete results appear in Appendix~\ref{sec:app-single-dfopd}.}
\label{fig:selfgen}
\end{figure}

We first evaluate DF-OPD in the two single-teacher settings. Each teacher generates its own training questions, and we examine pools of size $|D|\in\{1,8,64,1\mathrm{k}\}$. We compare these pools with OPD trained on the full DAPO dataset under matched training and evaluation protocols.

\textbf{Self-generated questions can replace external training data in both single-teacher settings.}
As shown in Figure~\ref{fig:selfgen}, just 64 generated questions yield performance comparable to full-data OPD. In the RL-tuned teacher setting, DF-OPD and full-data OPD score 56.61 and 56.41, respectively. In the larger teacher setting, the corresponding scores are 49.16 and 47.36, with DF-OPD even slightly outperforming the full-data baseline. These results show that, in both settings, the teacher can supply not only distillation supervision but also training questions that support effective distillation. Complete results appear in Appendix~\ref{sec:app-single-dfopd}.

\subsection{Extending DF-OPD to Multiple Teachers}
\label{sec:dfmopd}
\label{sec:mopd-results}

\begin{table}[t]
\centering
\caption{\textbf{Self-generated questions in multi-teacher distillation.} 
Each domain score averages its two benchmarks, and Overall is the macro average of the three domain scores. 
Recovery is relative to the routed teachers (Equation~\ref{eq:recovery}). 
All trained rows use the peak-Overall checkpoint. Full results appear in Appendix~\ref{sec:app-multi-dfopd}.}
\label{tab:mopd}
\setlength{\tabcolsep}{5pt}
\begin{tabular}{lrccccc}
\toprule
Input / reference & Questions & Math & Code & IF & Overall & Recovery \\
\midrule
Initial student & --- & 15.83 & 15.77 & 41.01 & 24.21 & --- \\
Routed teachers & --- & 23.02 & 22.52 & 48.73 & 31.42 & 100\% \\
\midrule
Real post-training data & 7,458 & 23.02 & 21.71 & 48.80 & 31.18 & 96.6\% \\
Self-generated & 8 & 23.44 & 19.64 & 45.92 & 29.66 & 75.6\% \\
Self-generated & 64 & 22.60 & 20.87 & 47.51 & 30.33 & 84.8\% \\
Self-generated & 256 & 23.54 & 20.99 & 47.86 & 30.80 & 91.3\% \\
Self-generated & 1,000 & 23.85 & 21.17 & 48.91 & \textbf{31.31} & \textbf{98.5\%} \\
\bottomrule
\end{tabular}
\end{table}

We further extend DF-OPD to the multi-teacher setting to examine whether self-generated questions can support capability transfer from multiple domain experts to a single student. Specifically, the mathematics, code, and instruction-following teachers each generate questions in their respective domains and supervise the student's trajectories on those questions.

We compare DF-OPD with a MOPD baseline trained on 7,458 (1,540/1,870/4,048) real post-training questions, where the counts in parentheses correspond to mathematics, code, and instruction following, respectively. The generated pools contain 8 (2/2/4), 64 (13/16/35), 256 (53/64/139), and $1\mathrm{k}$ (206/251/543) questions, preserving the real-data domain proportions as closely as possible.

In addition to the overall score, we report \emph{recovery}, which measures the fraction of the initial student--teacher performance gap recovered through distillation:
\begin{equation}
\label{eq:recovery}
\mathrm{Recovery}(S)=\frac{S-S_0}{T-S_0}\times100\%,
\end{equation}
where $S$ and $S_0$ are the overall scores of the distilled and initial students, respectively, and $T$ averages each domain teacher's score on its own domain. All scores use the same evaluation protocol.

\textbf{DF-OPD with 1k self-generated questions achieves overall performance comparable to real-data MOPD.}
As shown in Table~\ref{tab:mopd}, DF-OPD achieves an overall score of 31.31, compared with 31.18 for the real-data baseline, with recovery rates of 98.5\% and 96.6\%, respectively. Across domains, self-generated questions yield slightly higher scores in mathematics and instruction following and a slightly lower score in code. These results show that self-generated questions support both capability transfer from a single teacher and distillation from multiple domain experts into a single student.

\textbf{MOPD is more sensitive to the size of the generated question pool than single-teacher distillation.}
With 8, 64, 256, and $1\mathrm{k}$ generated questions, recovery reaches 75.6\%, 84.8\%, 91.3\%, and 98.5\%, respectively, showing clear gains at each increase in pool size. In contrast, single-teacher DF-OPD already matches the full-data baseline with 64 generated questions.

This difference may reflect two factors. First, the total question pool in MOPD is divided among three domains: a pool of 8 questions leaves only 2, 2, and 4 questions per domain, reducing the number of training questions available within each domain. Second, potential interference across domains~\citep{camopd,openmopd,mopd,uniopd} may place greater demands on within-domain signal coverage. A larger question pool may therefore help meet these coverage needs and better support joint distillation from multiple teachers. Complete benchmark results, domain allocations, and training curves appear in Appendix~\ref{sec:app-multi-dfopd}.

\textbf{Further Analyses of DF-OPD.}
Additional experiments show that 64 generated questions also support effective single-teacher distillation from each of the three domain experts, achieving most of the gains obtained with real-data training (Appendix~\ref{sec:app-dfopd-domain-experts}). We further assess sensitivity to topic guidance by removing explicit topic specifications from the question-generation prompts while keeping other settings unchanged. Across the two main single-teacher settings and MOPD, the resulting score changes range from $-0.60$ to $+1.50$ percentage points, suggesting that DF-OPD remains effective without explicit topic guidance and is relatively insensitive to this aspect of prompt design (Appendix~\ref{sec:app-dfopd-topic-sensitivity}).

\section{Can OPD Work Without Explicit Training Questions?}
\label{sec:template}

The preceding section shows that the system can supply its own training questions, eliminating the need for external data. We now ask whether OPD remains effective if explicit questions are removed.

\subsection{Distillation with Empty Prompts}
\label{sec:empty}

We construct empty prompts using the student's chat template, with no user or system prompt content, following~\citet{oneshotopd}, to examine whether OPD remains effective without explicit training questions. In both single-teacher settings, the student samples responses from its corresponding empty prompt and receives supervision from its teacher. Under MOPD, all domains use the same empty prompt, and domain labels route trajectories to the corresponding domain teachers. The proportions of mathematics, code, and instruction-following labels match those in the full dataset, at $20.6:25.1:54.3$, to avoid confounding the comparison with changes in domain proportions. Exact input templates are provided in Appendix~\ref{sec:app-template-curves}.

\begin{figure}[t]
\centering
\includegraphics[width=\linewidth]{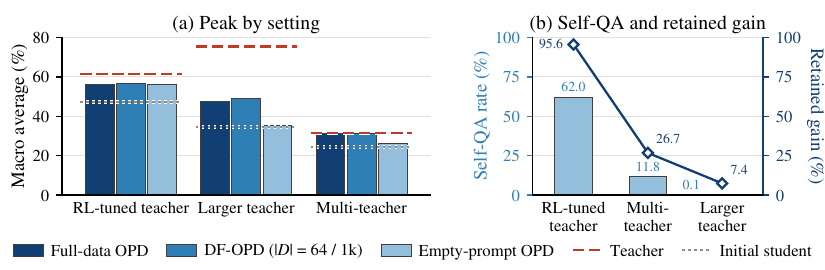}
\caption{\textbf{Omitting questions has a setting-dependent effect.} (a) Peak scores: six-benchmark mathematics averages (single-teacher; DF-OPD with 64 questions) or three-domain averages (MOPD; DF-OPD with 1k). (b) Self-QA rate of each initial student (1,024 empty-prompt rollouts, setting-specific domain criteria; Appendix~\ref{sec:app-selfqa}) and the full-data gain retained by empty-prompt OPD.}
\label{fig:empty-prompt}
\end{figure}

\textbf{Empty-prompt OPD retains most of the full-data gains only in some settings.}
As shown in Figure~\ref{fig:empty-prompt}, empty-prompt OPD reaches 56.00 in the RL-tuned teacher setting, close to full-data OPD and DF-OPD, retaining approximately 95.6\% of the gain achieved by full-data OPD over the initial student. However, its scores reach only 35.44 and 26.07 in the larger teacher and multi-teacher settings, retaining only approximately 7.4\% and 26.7\% of the corresponding full-data gains, respectively.

\subsection{Empty-Prompt Rollouts as Implicit DF-OPD}
\label{sec:implicit}

To understand this difference, we inspect student behavior under empty prompts and observe an unexpected phenomenon in some models: \textbf{self-questioning and answering (Self-QA)}. Without receiving a user question, these models pose a question themselves and then attempt to answer it within the same rollout.

To quantify this behavior, we sample 1,024 empty-prompt rollouts from each student before training and measure the fraction exhibiting Self-QA. The Self-QA rate reaches 62.0\% for the R1-Distill student, compared with 11.8\% for SmolLM3 and 0.1\% for Qwen3. Across these students, higher Self-QA rates coincide with greater retention of full-data gains: the three students retain 95.6\%, 26.7\%, and 7.4\%, respectively (Figure~\ref{fig:empty-prompt}(b)). Sampling and classification details are in Appendix~\ref{sec:app-selfqa}.

Appendix~\ref{sec:app-rollouts} presents representative examples: R1-Distill poses an asymptote problem and attempts to solve it under an empty prompt, whereas Qwen3 asks the user to provide a question. In such Self-QA rollouts, the student's own question supplies task content for the subsequent solution attempt and elicits teacher supervision. This process can be viewed as \textbf{implicit DF-OPD}, with question generation embedded in the student's sampling process. 

Whereas Self-QA arises inconsistently across students, DF-OPD uses teachers to explicitly generate questions, allowing control over question counts, domains, and their mixture without relying on spontaneous student behavior.
These results show that, in the settings studied, OPD can dispense with external training data and achieve performance comparable to full-data OPD using only questions generated and supervision provided by the system itself. Further removing explicit questions, as with empty prompts, does not work as consistently across settings.

\section{Related Work}
\label{sec:related}

Existing work on OPD studies supervision and optimization~\citep{opdsurvey}, token importance~\citep{tip}, and alignment over high-probability token supports~\citep{rethinkingopd}. Data-centric work investigates difficulty-aware balancing~\citep{uniopd} and reliability-aware prompt ordering~\citep{reorderopd}. 
Concurrent studies also reveal OPD's limited sensitivity to training data~\citep{oneshotopd,whatmattersopd,everyconin}, but the scope and limits of this insensitivity remain underexplored.
Our work takes this line of inquiry further by systematically investigating \emph{how far can OPD go without external data?}.
We first broaden the analysis across three dimensions of training data---quantity, dataset-level difficulty, and initial distillation gap---and examine the underlying phenomena through supervision signal coverage and cross-domain reasoning transfer.
We then move beyond characterizing this insensitivity to eliminating dependence on external question corpora: we introduce DF-OPD, in which the system supplies its own training questions, and demonstrate its effectiveness in single-teacher and multi-teacher settings across multiple task domains.
Finally, we investigate whether OPD can dispense with explicit training questions altogether~\citep{oneshotopd}, revealing the setting-dependent effectiveness of empty-prompt distillation and its connection to implicit DF-OPD through student self-questioning and answering.
Beyond OPD, small datasets also support alignment and reasoning through SFT~\citep{lima,limo,s1} and RL~\citep{limr,oneshotrlvr}. Additional related work on synthetic data and multi-teacher distillation appears in Appendix~\ref{sec:app-related}.

\section{Conclusion}
\label{sec:conclusion}

We examined how far OPD's dependence on training questions can be reduced. Across two single-teacher settings, few questions and datasets with different difficulty and initial distillation gaps yield comparable outcomes. 
Signal-coverage and cross-domain analyses suggest that repeated sampling exposes substantial supervision and that transferred capabilities extend beyond dataset-specific knowledge. 
Self-generated questions also support effective single- and multi-teacher distillation without external question corpora. 
Omitting explicit questions exposes a boundary: success varies across the examined settings and can coincide with student self-questioning, an implicit form of DF-OPD. 
Our experiments cover a limited range of teacher--student combinations, model scales, task domains, and training budgets, and the number of questions required may vary with these factors; future work should evaluate larger models, broader tasks, and longer training. Moreover, question efficiency does not imply computational efficiency, as OPD still requires repeated student rollouts and teacher feedback; improving computational efficiency is another direction for future work. 
Together, these results invite a reassessment of the role of training data in on-policy distillation.

\label{conclusion:end}
\label{maintext:end}
\subsection*{AI use statement}
We used generative AI tools to assist with programming and debugging, and to polish the writing of the manuscript for readability. The authors reviewed all AI-assisted content, such as code and edited text, and take responsibility for the final content of this work, including text, claims, or artifacts produced with the aid of generative AI. The teacher-generated training questions in DF-OPD (Section~\ref{sec:dfopd}) are part of the studied method rather than assistance in preparing this work.

\subsection*{Ethics statement}
This work aims to understand and reduce the training-data requirements of on-policy distillation. Our experiments use existing models and datasets together with teacher-generated questions, focusing on mathematics, programming, and instruction following. Reducing reliance on external training-question corpora may lower the burden of data collection and curation and make distillation research more accessible. We encourage responsible use of the resulting models in accordance with applicable usage terms and appropriate evaluation practices.

\subsection*{Reproducibility statement}
We are committed to the reproducibility of our results. All teacher and student models are publicly released checkpoints, and all external training questions and benchmarks come from public sources. Code for DF-OPD is available at the repository linked in the abstract, and the complete code will be open-sourced upon acceptance.

\bibliography{iclr2027_conference}

@inproceedings{minillm,
title={Mini{LLM}: Knowledge Distillation of Large Language Models},
author={Yuxian Gu and Li Dong and Furu Wei and Minlie Huang},
booktitle={The Twelfth International Conference on Learning Representations},
year={2024},
url={https://openreview.net/forum?id=5h0qf7IBZZ}
}

@article{exopd,
  title={Learning beyond teacher: Generalized on-policy distillation with reward extrapolation},
  author={Yang, Wenkai and Liu, Weijie and Xie, Ruobing and Yang, Kai and Yang, Saiyong and Lin, Yankai},
  journal={arXiv preprint arXiv:2602.12125},
  year={2026}
}

@inproceedings{
rethinkingopd,
title={Rethinking On-Policy Distillation of Large Language Models: Phenomenology, Mechanism, and Recipe},
author={Yaxuan Li and Yuxin Zuo and Bingxiang He and Jinqian Zhang and Chaojun Xiao and Cheng Qian and Tianyu Yu and {Huan-ang} Gao and Wenkai Yang and Zhiyuan Liu and Ning Ding},
booktitle={ICML 2026 Workshop on Foundations of Deep Generative Models: Understanding Memorization, Generalization, and Reasoning},
year={2026},
url={https://openreview.net/forum?id=aaDDXXDGXB}
}

@inproceedings{gkd,
title={On-Policy Distillation of Language Models: Learning from Self-Generated Mistakes},
author={Rishabh Agarwal and Nino Vieillard and Yongchao Zhou and Piotr Stanczyk and Sabela Ramos Garea and Matthieu Geist and Olivier Bachem},
booktitle={The Twelfth International Conference on Learning Representations},
year={2024},
url={https://openreview.net/forum?id=3zKtaqxLhW}
}

@article{deepseekr1,
  title={Deepseek-r1: Incentivizing reasoning capability in llms via reinforcement learning},
  author={Guo, Daya and Yang, Dejian and Zhang, Haowei and Song, Junxiao and Wang, Peiyi and Zhu, Qihao and Xu, Runxin and Zhang, Ruoyu and Ma, Shirong and Bi, Xiao and others},
  journal={arXiv preprint arXiv:2501.12948},
  year={2025}
}

@article{qwen3,
  title={Qwen3 technical report},
  author={Yang, An and Li, Anfeng and Yang, Baosong and Zhang, Beichen and Hui, Binyuan and Zheng, Bo and Yu, Bowen and Gao, Chang and Huang, Chengen and Lv, Chenxu and others},
  journal={arXiv preprint arXiv:2505.09388},
  year={2025}
}

@inproceedings{
revisitingopd,
title={Revisiting On-Policy Distillation: Empirical Failure Modes and Simple Fixes},
author={Yuqian Fu and Haohuan Huang and Kaiwen Jiang and Jiacai Liu and Zhuo Jiang and Yuanheng Zhu and Dongbin Zhao},
booktitle={Third Conference on Language Modeling},
year={2026},
url={https://openreview.net/forum?id=uZbZb058bA}
}

@article{opdsurvey,
  title={A survey of on-policy distillation for large language models},
  author={Song, Mingyang and Zheng, Mao},
  journal={arXiv preprint arXiv:2604.00626},
  year={2026}
}

@inproceedings{
eopd,
title={Entropy-Aware On-Policy Distillation of Language Models},
author={Woogyeol Jin and Taywon Min and Yongjin Yang and Dennis Wei and Yi Zhou and Swanand Ravindra Kadhe and Nathalie Baracaldo and Kimin Lee},
booktitle={Forty-third International Conference on Machine Learning},
year={2026},
url={https://openreview.net/forum?id=J5i09faOOf}
}

@article{tip,
  title={Tip: Token importance in on-policy distillation},
  author={Xu, Yuanda and Sang, Hejian and Zhou, Zhengze and He, Ran and Wang, Zhipeng and Geramifard, Alborz},
  journal={arXiv preprint arXiv:2604.14084},
  year={2026}
}

@article{thinkingmachineslab,
  author = {Kevin Lu and {Thinking Machines Lab}},
  title = {On-Policy Distillation},
  journal = {Thinking Machines Lab: Connectionism},
  year = {2025},
  note = {https://thinkingmachines.ai/blog/on-policy-distillation},
  doi = {10.64434/tml.20251026},
}

@article{mimov2flash,
  title={Mimo-v2-flash technical report},
  author={Xiao, Bangjun and Xia, Bingquan and Yang, Bo and Gao, Bofei and Shen, Bowen and Zhang, Chen and He, Chenhong and Lou, Chiheng and Luo, Fuli and Wang, Gang and others},
  journal={arXiv preprint arXiv:2601.02780},
  year={2026}
}

@article{uniopd,
  title={Uni-opd: Unifying on-policy distillation with a dual-perspective recipe},
  author={Hou, Wenjin and Peng, Shangpin and Wang, Weinong and Ruan, Zheng and Zhang, Yue and Zhou, Zhenglin and Gao, Mingqi and Chen, Yifei and Wang, Kaiqi and Yang, Hongming and others},
  journal={arXiv preprint arXiv:2605.03677},
  year={2026}
}

@article{deepseekv4,
  title={Deepseek-v4: Towards highly efficient million-token context intelligence},
  author={Xu, Anyi and Lin, Bangcai and Xue, Bing and Wang, Bingxuan and Xu, Bingzheng and Wu, Bochao and Zhang, Bowei and Lin, Chaofan and Dong, Chen and Ling, Chenchen and others},
  journal={arXiv preprint arXiv:2606.19348},
  year={2026}
}

@article{opdeltad,
  title={On-Policy Delta Distillation},
  author={Heo, Byeongho and Hwang, Jaehui and Yun, Sangdoo and Han, Dongyoon},
  journal={arXiv preprint arXiv:2607.15161},
  year={2026}
}

@article{trustopd,
  title={Trust Region On-Policy Distillation},
  author={Xing, Xingrun and Wang, Haoqing and Gao, Boyan and Li, Ziheng and Tang, Yehui},
  journal={arXiv preprint arXiv:2606.01249},
  year={2026}
}

@article{trustpd,
  title={Trust Region Policy Distillation},
  author={Xie, Zhengpeng and Zhang, Li Lyna and Xie, Zeke and Yang, Mao},
  journal={arXiv preprint arXiv:2607.04751},
  year={2026}
}

@article{justrl,
  title={Justrl: Scaling a 1.5 b llm with a simple rl recipe},
  author={He, Bingxiang and Qu, Zekai and Liu, Zeyuan and Chen, Yinghao and Zuo, Yuxin and Qian, Cheng and Zhang, Kaiyan and Chen, Weize and Xiao, Chaojun and Cui, Ganqu and others},
  journal={arXiv preprint arXiv:2512.16649},
  year={2025}
}

@inproceedings{deepmath,
  title={Deepmath-103k: A large-scale, challenging, decontaminated, and verifiable mathematical dataset for advancing reasoning},
  author={He, Zhiwei and Liang, Tian and Xu, Jiahao and Liu, Qiuzhi and Chen, Xingyu and Wang, Yue and Song, Linfeng and Yu, Dian and Liang, Zhenwen and Wang, Wenxuan and others},
  booktitle={International Conference on Learning Representations},
  volume={2026},
  pages={138306--138322},
  year={2026}
}

@article{dapo,
  title={Dapo: An open-source llm reinforcement learning system at scale},
  author={Yu, Qiying and Zhang, Zheng and Zhu, Ruofei and Yuan, Yufeng and Zuo, Xiaochen and Yue, Yu and Dai, Weinan and Fan, Tiantian and Liu, Gaohong and Liu, Lingjun and others},
  journal={Advances in Neural Information Processing Systems},
  volume={38},
  pages={113222--113244},
  year={2025}
}

@inproceedings{verl,
  title={Hybridflow: A flexible and efficient rlhf framework},
  author={Sheng, Guangming and Zhang, Chi and Ye, Zilingfeng and Wu, Xibin and Zhang, Wang and Zhang, Ru and Peng, Yanghua and Lin, Haibin and Wu, Chuan},
  booktitle={Proceedings of the Twentieth European Conference on Computer Systems},
  pages={1279--1297},
  year={2025}
}

@article{openmopd,
  title={Open-MOPD: Diagnosing and Fixing Capability Imbalance in Multi-Teacher On-Policy Distillation},
  author={Gao, Huan-ang and Chi, Haohan and Yan, Yong and Feng, Shiyuan and Wu, Hanlin and Jiang, Zheng and He, Bingxiang and Ma, Wei-Ying and Zhang, Ya-Qin and Zhou, Hao},
  journal={arXiv preprint arXiv:2608.19098},
  year={2026}
}

@misc{aime2024,
  author = {AI-MO},
  title = {AIME 2024},
  year = {2024},
  howpublished = {\url{https://huggingface.co/datasets/AI-MO/aimo-validation-aime}}
}

@misc{amc2023,
  author = {AI-MO},
  title = {AMC 2023},
  year = {2024},
  howpublished = {\url{https://huggingface.co/datasets/AI-MO/aimo-validation-amc}}
}

@misc{aime2025,
  author = {OpenCompass},
  title = {AIME 2025},
  year = {2025},
  howpublished = {\url{https://huggingface.co/datasets/opencompass/AIME2025}}
}

@misc{aime26,
      title={American Invitational Mathematics Examination (AIME) 2026}, 
      author={Zhang, Yifan and {Math-AI Team}},
      year={2026},
      howpublished = {\url{https://huggingface.co/datasets/math-ai/aime26}}
}

@inproceedings{livecodebench,
title={LiveCodeBench: Holistic and Contamination Free Evaluation of Large Language Models for Code},
author={Naman Jain and King Han and Alex Gu and Wen-Ding Li and Fanjia Yan and Tianjun Zhang and Sida Wang and Armando Solar-Lezama and Koushik Sen and Ion Stoica},
booktitle={The Thirteenth International Conference on Learning Representations},
year={2025},
url={https://openreview.net/forum?id=chfJJYC3iL}
}

@article{ifeval,
  title={Instruction-following evaluation for large language models},
  author={Zhou, Jeffrey and Lu, Tianjian and Mishra, Swaroop and Brahma, Siddhartha and Basu, Sujoy and Luan, Yi and Zhou, Denny and Hou, Le},
  journal={arXiv preprint arXiv:2311.07911},
  year={2023}
}

@article{ifbench,
  title={Generalizing verifiable instruction following},
  author={Pyatkin, Valentina and Malik, Saumya and Graf, Victoria and Ivison, Hamish and Huang, Shengyi and Dasigi, Pradeep and Lambert, Nathan and Hajishirzi, Hanna},
  journal={Advances in Neural Information Processing Systems},
  volume={38},
  year={2025}
}

@article{orz,
  title={Open-reasoner-zero: An open source approach to scaling up reinforcement learning on the base model},
  author={Hu, Jingcheng and Zhang, Yinmin and Han, Qi and Jiang, Daxin and Zhang, Xiangyu and Shum, Heung-Yeung},
  journal={Advances in Neural Information Processing Systems},
  volume={38},
  pages={162239--162262},
  year={2025}
}

@misc{deepcoder,
  title={DeepCoder: A Fully Open-Source 14B Coder at O3-mini Level},
  author={Michael Luo and Sijun Tan and Roy Huang and Ameen Patel and Alpay Ariyak and Qingyang Wu and Xiaoxiang Shi and Rachel Xin and Colin Cai and Maurice Weber and Ce Zhang and Li Erran Li and Raluca Ada Popa and Ion Stoica},
  howpublished={\url{https://pretty-radio-b75.notion.site/DeepCoder-A-Fully-Open-Source-14B-Coder-at-O3-mini-Level-1cf81902c14680b3bee5eb349a512a51}},
  note={Notion Blog},
  year={2025}
}

@article{mathdataset,
  title={Measuring mathematical problem solving with the math dataset},
  author={Hendrycks, Dan and Burns, Collin and Kadavath, Saurav and Arora, Akul and Basart, Steven and Tang, Eric and Song, Dawn and Steinhardt, Jacob},
  journal={arXiv preprint arXiv:2103.03874},
  year={2021}
}

@inproceedings{math500,
  title={Let's verify step by step},
  author={Lightman, Hunter and Kosaraju, Vineet and Burda, Yuri and Edwards, Harrison and Baker, Bowen and Lee, Teddy and Leike, Jan and Schulman, John and Sutskever, Ilya and Cobbe, Karl},
  booktitle={International Conference on Learning Representations},
  volume={2024},
  pages={39578--39601},
  year={2024}
}

@article{minerva,
  title={Solving quantitative reasoning problems with language models},
  author={Lewkowycz, Aitor and Andreassen, Anders and Dohan, David and Dyer, Ethan and Michalewski, Henryk and Ramasesh, Vinay and Slone, Ambrose and Anil, Cem and Schlag, Imanol and Gutman-Solo, Theo and others},
  journal={Advances in neural information processing systems},
  volume={35},
  pages={3843--3857},
  year={2022}
}

@inproceedings{olympiadbench,
  title={Olympiadbench: A challenging benchmark for promoting agi with olympiad-level bilingual multimodal scientific problems},
  author={He, Chaoqun and Luo, Renjie and Bai, Yuzhuo and Hu, Shengding and Thai, Zhen and Shen, Junhao and Hu, Jinyi and Han, Xu and Huang, Yujie and Zhang, Yuxiang and others},
  booktitle={Proceedings of the 62nd Annual Meeting of the Association for Computational Linguistics (Volume 1: Long Papers)},
  pages={3828--3850},
  year={2024}
}

@article{camopd,
  title={Counteraction-Aware Multi-Teacher On-Policy Distillation for General Capability Recovery with Domain Preservation},
  author={Chen, Tianlei and Ou, Jiao and Liu, Ziyuan and Tang, Ruiming and Liang, Jian and Li, Han},
  journal={arXiv preprint arXiv:2605.27115},
  year={2026}
}

@inproceedings{rzero,
  title={R-zero: Self-evolving reasoning llm from zero data},
  author={Huang, Chengsong and Yu, Wenhao and Wang, Xiaoyang and Zhang, Hongming and Li, Zongxia and Li, Ruosen and Huang, Jiaxin and Mi, Haitao and Yu, Dong},
  booktitle={International Conference on Learning Representations},
  volume={2026},
  pages={130770--130790},
  year={2026}
}

@article{rdiverse,
  title={R-diverse: Mitigating diversity illusion in self-play llm training},
  author={Li, Gengsheng and He, Jinghan and Wang, Shijie and Zhang, Dan and Liu, Ruiqi and Zhang, Renrui and Yao, Zijun and Fang, Junfeng and Guo, Haiyun and Wang, Jinqiao},
  journal={arXiv preprint arXiv:2602.13103},
  year={2026}
}

@article{nc2,
  title={Nemotron-cascade 2: Post-training llms with cascade rl and multi-domain on-policy distillation},
  author={Yang, Zhuolin and Liu, Zihan and Chen, Yang and Dai, Wenliang and Wang, Boxin and Lin, Sheng-Chieh and Lee, Chankyu and Chen, Yangyi and Jiang, Dongfu and He, Jiafan and others},
  journal={arXiv preprint arXiv:2603.19220},
  year={2026}
}

@article{oneshotrlvr,
  title={Reinforcement learning for reasoning in large language models with one training example},
  author={Wang, Yiping and Yang, Qing and Zeng, Zhiyuan and Ren, Liliang and Liu, Liyuan and Peng, Baolin and Cheng, Hao and He, Xuehai and Wang, Kuan and Gao, Jianfeng and others},
  journal={Advances in Neural Information Processing Systems},
  volume={38},
  pages={122721--122764},
  year={2025}
}

@article{oneshotopd,
  title={Rethinking On-Policy Distillation of Large Language Models II: One Training Example},
  author={Fu, Zixuan and He, Bingxiang and Zuo, Yuxin and Huang, Haohuan and Zhang, Jinqian and Xiao, Ruhang and Qian, Cheng and Luo, Qinyu and Gao, Huan-ang and Wang, Yudong and others},
  journal={arXiv preprint arXiv:2609.04172},
  year={2026}
}

@article{whatmattersopd,
  title={What Matters in On-Policy Distillation? A Perspective on Data Efficiency and Data Selection},
  author={Hou, Zhinan and Zhang, Jiaqi and Cai, Xunliang and You, Keyou},
  journal={arXiv preprint arXiv:2609.05198},
  year={2026}
}

@article{reorderopd,
  title={ReOrder-OPD: Reliability-Aware Prompt Ordering for On-Policy Distillation},
  author={Zhu, Ximo and Liu, Ruiqi and Wang, Rong and Wu, Ping and Zheng, Xiang and Xu, Wenzhuo and Yao, Xubin and Yan, Zhiyuan and Li, Bo and Gao, Jun and others},
  journal={arXiv preprint arXiv:2608.10905},
  year={2026}
}

@article{lima,
  title={Lima: Less is more for alignment},
  author={Zhou, Chunting and Liu, Pengfei and Xu, Puxin and Iyer, Srinivasan and Sun, Jiao and Mao, Yuning and Ma, Xuezhe and Efrat, Avia and Yu, Ping and Yu, Lili and others},
  journal={Advances in Neural Information Processing Systems},
  volume={36},
  pages={55006--55021},
  year={2023}
}

@article{limo,
  title={Limo: Less is more for reasoning},
  author={Ye, Yixin and Huang, Zhen and Xiao, Yang and Chern, Ethan and Xia, Shijie and Liu, Pengfei},
  journal={arXiv preprint arXiv:2502.03387},
  year={2025}
}

@inproceedings{s1,
  title={s1: Simple test-time scaling},
  author={Muennighoff, Niklas and Yang, Zitong and Shi, Weijia and Li, Xiang Lisa and Fei-Fei, Li and Hajishirzi, Hannaneh and Zettlemoyer, Luke and Liang, Percy and Cand{\`e}s, Emmanuel and Hashimoto, Tatsunori B},
  booktitle={Proceedings of the 2025 Conference on Empirical Methods in Natural Language Processing},
  pages={20286--20332},
  year={2025}
}

@article{limr,
  title={Limr: Less is more for rl scaling},
  author={Li, Xuefeng and Zou, Haoyang and Liu, Pengfei},
  journal={arXiv preprint arXiv:2502.11886},
  year={2025}
}

@article{mopd,
  title={Mopd: Multi-teacher on-policy distillation for capability integration in llm post-training},
  author={Ma, Wenhan and Wei, Jianyu and Zhao, Liang and Zhang, Hailin and Xiao, Bangjun and Li, Lei and Yang, Qibin and Gao, Bofei and Wang, Yudong and Li, Rang and others},
  journal={arXiv preprint arXiv:2606.30406},
  year={2026}
}

@inproceedings{selfinstruct,
  title={Self-instruct: Aligning language models with self-generated instructions},
  author={Wang, Yizhong and Kordi, Yeganeh and Mishra, Swaroop and Liu, Alisa and Smith, Noah A and Khashabi, Daniel and Hajishirzi, Hannaneh},
  booktitle={Proceedings of the 61st annual meeting of the association for computational linguistics (volume 1: long papers)},
  pages={13484--13508},
  year={2023}
}

@inproceedings{evolinstruct,
  title={WizardLM: Empowering large pre-trained language models to follow complex instructions},
  author={Xu, Can and Sun, Qingfeng and Zheng, Kai and Geng, Xiubo and Zhao, Pu and Feng, Jiazhan and Tao, Chongyang and Lin, Qingwei and Jiang, Daxin},
  booktitle={International Conference on Learning Representations},
  volume={2024},
  pages={30745--30766},
  year={2024}
}

@inproceedings{magpie,
  title={Magpie: Alignment data synthesis from scratch by prompting aligned llms with nothing},
  author={Xu, Zhangchen and Jiang, Fengqing and Niu, Luyao and Deng, Yuntian and Poovendran, Radha and Choi, Yejin and Lin, Bill Yuchen},
  booktitle={International Conference on Learning Representations},
  volume={2025},
  pages={76346--76382},
  year={2025}
}

@inproceedings{absolutezero,
title={Absolute Zero: Reinforced Self-play Reasoning with Zero Data},
author={Andrew Zhao and Yiran Wu and Yang Yue and Tong Wu and Quentin Xu and Yang Yue and Matthieu Lin and Shenzhi Wang and Qingyun Wu and Zilong Zheng and Gao Huang},
booktitle={The Thirty-ninth Annual Conference on Neural Information Processing Systems},
year={2025},
url={https://openreview.net/forum?id=neZSGqhxDa}
}

@article{adamw,
  title={Decoupled weight decay regularization},
  author={Loshchilov, Ilya and Hutter, Frank},
  journal={arXiv preprint arXiv:1711.05101},
  year={2017}
}

@article{trustmopd,
  title={Distill What You Trust: Reliability-Aware Multi-Teacher On-Policy Distillation},
  author={Sun, Jie and Zheng, Mao and Song, Mingyang and Liu, Zeyuan and Li, Gengsheng and Jiang, Houcheng and Cheng, Yilin and Feng, Bichuan and Cai, Yuchen and Fang, Junfeng and others},
  journal={arXiv preprint arXiv:2609.23697},
  year={2026}
}

@article{everyconin,
  title={Every Coin Has Two Sides: On the Dual Nature of Generalization in On-Policy Distillation of Large Language Models},
  author={Li, Zhaoyi and Kong, Deyang and Wei, Yuan and Yang, Evan and Shen, Ranran and Ihsani, Mahardika Krisna and Yang, Ming and Zhang, Wei and Hao, Chuan and Yang, Jian and others},
  journal={arXiv preprint arXiv:2608.16647},
  year={2026}
}
\bibliographystyle{iclr2027_conference}

\appendix
\section{OPD Training Objective}
\label{sec:app-opd}

The main text presents the full-vocabulary reverse KL objective. In practice, we use a stop-gradient surrogate objective over the student's top-$k$ support to provide token-level supervision at each state. Let $\theta_j$ denote the student parameters used to generate the current batch of trajectories. Given a question $x$, we sample a response $y\sim\pi_{\theta_j}(\cdot\mid x)$ and denote the state at response position $t$ by $s_t=(x,y_{<t})$. Let
\begin{equation}
p_j(v\mid s)=\pi_{\theta_j}(v\mid s),
\qquad
q(v\mid s)=\pi_\phi(v\mid s).
\end{equation}

For each response state $s_t$, we define the student's top-$k$ support, the normalized probabilities within this support, and the teacher--student log-probability difference as
\begin{align}
V_t
&=
\operatorname{TopK}_k\!\left(p_j(\cdot\mid s_t)\right),
\\
\widetilde p_j(v\mid s_t)
&=
\frac{p_j(v\mid s_t)}
{\sum_{u\in V_t}p_j(u\mid s_t)},
\qquad v\in V_t,
\\
\delta_t(v)
&=
\log q(v\mid s_t)-\log p_j(v\mid s_t).
\end{align}
Both the support and its normalized weights are determined by the student distribution. The two log-probabilities in $\delta_t$ are normalized over the full vocabulary, whereas the multiplicative weights $\widetilde p_j$ are renormalized within $V_t$.

The training loss at each response state is
\begin{equation}
\ell_t(\theta)
=
-\sum_{v\in V_t}
\operatorname{sg}\!\left[
\widetilde p_j(v\mid s_t)\delta_t(v)
\right]
\log\pi_\theta(v\mid s_t),
\end{equation}
where $\operatorname{sg}$ denotes the stop-gradient operator. During the update, the support sets, sampled trajectories, and weights inside the brackets are held fixed, while gradients flow through the student's log-probabilities. The implementation first averages the loss over valid response tokens within each microbatch, then accumulates gradients with each microbatch weighted by its share of the total sequence count on the corresponding process, and finally averages gradients across distributed processes.

\paragraph{Teacher routing.}
In MOPD, the domain label $d(x)$ of each question selects a single teacher for the entire response:
\begin{equation}
q(v\mid s)
=
\pi_{\phi_{d(x)}}(v\mid s).
\end{equation}
All domains use the same training objective, with each domain's contribution determined by the sampled responses and their token-level supervision.

\FloatBarrier
\section{Experimental Setup}
\label{sec:app-experimental}

\subsection{Models}
\label{sec:app-models}

Table~\ref{tab:app-model-pairs} lists the five teacher--student
relationships used in the two single-teacher settings and the
three-domain multi-teacher setting. All models are initialized from
eight publicly released checkpoints; their original supervised
fine-tuning and reinforcement learning were performed by the model
providers.

\begin{table}[htbp]
\centering
\caption{
\textbf{Teacher and student configurations.} Model names refer to publicly
released checkpoints. The three MOPD domains share the same student
initialization and use the checkpoint collection released by Open-MOPD.
}
\label{tab:app-model-pairs}
\small
\setlength{\tabcolsep}{4pt}
\begin{tabular}{
@{}
>{\raggedright\arraybackslash}p{0.14\linewidth}
>{\raggedright\arraybackslash}p{0.27\linewidth}
>{\raggedright\arraybackslash}p{0.29\linewidth}
>{\raggedright\arraybackslash}p{0.23\linewidth}
@{}
}
\toprule
Setting & Student & Teacher & Relationship \\
\midrule
RL-tuned teacher
& DeepSeek-R1-Distill-Qwen-1.5B
& JustRL-DeepSeek-1.5B
& RL-tuned from the student \\
Larger teacher
& Qwen3-1.7B
& Qwen3-30B-A3B-Instruct-2507
& Different-scale models from the Qwen3 family \\
MOPD: math
& SmolLM3-3B-MixSFT
& SmolLM3-3B-RL-Math
& Math-specific RL from MixSFT \\
MOPD: code
& SmolLM3-3B-MixSFT
& SmolLM3-3B-RL-Code
& Code-specific RL from MixSFT \\
MOPD: instruction following
& SmolLM3-3B-MixSFT
& SmolLM3-3B-RL-IF
& Instruction-following RL from MixSFT \\
\bottomrule
\end{tabular}
\end{table}

\paragraph{Teacher--student relationships.}
JustRL-DeepSeek-1.5B is obtained by GRPO post-training of
DeepSeek-R1-Distill-Qwen-1.5B.
The larger Qwen3 teacher is a mixture-of-experts model with
approximately 30B total parameters and 3B activated parameters;
it is not obtained by RL-tuning the 1.7B student.
In Open-MOPD, MixSFT is obtained by mixed-domain supervised
fine-tuning of SmolLM3-3B-Base.
The model provider separately applies domain-specific RL to MixSFT
to obtain the three domain teachers, and our MOPD student is
initialized from the same MixSFT checkpoint.

\subsection{Datasets}
\label{sec:app-datasets}

\paragraph{Single-teacher settings.}
For mathematics data, we use DAPO-Math-17k as released,
containing 17,917 records.
The prepared DeepMath question pool contains 14,116 records.
ORZ preprocessing includes deduplication, removal of identified
records overlapping with evaluation benchmarks, answer-field
cleanup, and length filtering, yielding 52,738 records.

For out-of-domain data, we preprocess DeepCoder and further filter
out mathematics-related problems.
We first exclude function-signature-style problems and apply
deduplication and length filtering, yielding 12,492 problems.
We then construct a variant with mathematics-related content removed:
we retain only TACO problems that have topic annotations but no
mathematics-related topic tags, and remove problems that an LLM judge
classifies as essentially mathematical or that exceed the prompt
length limit. This yields a final pool of 2,848 problems.

\paragraph{Multi-teacher setting.}
Following the data construction procedure of
TrustMOPD~\citep{trustmopd}, we sample from the
\texttt{rl\_prompt\_mix} data released by Open-MOPD using random
seed 42.
The resulting pool contains 1,540 mathematics records,
1,870 code records, and 4,048 instruction-following records,
for a total of 7,458 real training examples.
Distillation uses the question text and, in MOPD, the domain label
for teacher routing.

\paragraph{Few-shot question pools.}
For all few-shot experiments, we randomly sample the required number
of questions from the corresponding dataset.
For MOPD, we approximately preserve the mathematics, code, and
instruction-following proportions of the real-data pool:
20.6\%, 25.1\%, and 54.3\%, respectively.

\subsection{Training and Evaluation Protocol}
\label{sec:app-training}

\paragraph{Training configuration.}
In the single-teacher settings, we follow
Rethinking-OPD~\citep{rethinkingopd} and set the student support
size to $k=16$.
In the multi-teacher setting, we follow
TrustMOPD~\citep{trustmopd} and use $k=128$.
Table~\ref{tab:app-training} provides the training configuration.
Objective normalization and teacher routing are described in
Appendix~\ref{sec:app-opd}.

\begin{table}[htbp]
\centering
\caption{
\textbf{Training configurations for the single-teacher and multi-teacher
settings.} The student support size $k$ is used to compute the
distillation objective and is distinct from decoding top-$k$
during generation; decoding top-$k$ is disabled in both settings.
}
\label{tab:app-training}
\small
\setlength{\tabcolsep}{5pt}
\begin{tabular}{@{}lll@{}}
\toprule
Configuration & Single teacher & Multiple teachers \\
\midrule
Questions per update & 64 & 64 \\
Student support size $k$ & 16 & 128 \\
Policy passes per rollout batch & 1 & 1 \\
Optimizer & AdamW & AdamW \\
Learning rate & $10^{-6}$ & $10^{-6}$ \\
AdamW $\beta_1,\beta_2$ & 0.9, 0.999 & 0.9, 0.999 \\
Weight decay & 0.01 & 0.01 \\
Learning-rate schedule & Constant; no warmup & Constant; no warmup \\
Gradient-norm clipping & 1.0 & 1.0 \\
Sampling temperature / top-$p$ & 1.0 / 1.0 & 1.0 / 1.0 \\
Decoding top-$k$ & Disabled ($-1$) & Disabled ($-1$) \\
Actor dtype & FP32 & FP32 \\
Rollout dtype & BF16 & BF16 \\
Parameter sharding / rollout TP & FSDP / 2 & FSDP / 2 \\
\bottomrule
\end{tabular}
\end{table}

\paragraph{Evaluation configuration.}
Single-teacher mathematics evaluation uses $\mathrm{avg}@4$,
with a sampling temperature of 0.7 following
Rethinking-OPD~\citep{rethinkingopd}.
For MOPD, mathematics and code evaluation use $\mathrm{avg}@16$
and $\mathrm{avg}@5$, respectively, while instruction-following
evaluation uses one response per problem.
Following TrustMOPD~\citep{trustmopd}, we use a sampling temperature
of 0.6 for all three domains.
Table~\ref{tab:app-evaluation} provides the evaluation configuration.

\begin{table}[htbp]
\centering
\caption{
\textbf{Evaluation configurations for the single-teacher and multi-teacher
settings.} Responses per problem denotes the number of independently
sampled responses for each evaluation problem.
}
\label{tab:app-evaluation}
\small
\setlength{\tabcolsep}{4pt}
\begin{tabular}{@{}lcccc@{}}
\toprule
Configuration
& Single-teacher math
& MOPD math
& MOPD code
& MOPD IF \\
\midrule
Responses per problem & 4 & 16 & 5 & 1 \\
Sampling temperature & 0.7 & 0.6 & 0.6 & 0.6 \\
Top-$p$ & 0.95 & 0.95 & 0.95 & 0.95 \\
Response limit (tokens) & 16,384 & 32,768 & 32,768 & 32,768 \\
\bottomrule
\end{tabular}
\end{table}

\paragraph{Checkpoint selection.}
We save checkpoints every 15 updates in the two single-teacher
settings and every 20 updates in MOPD, and evaluate the saved
checkpoints.
For each experiment, we report the highest macro-average score
across all evaluated checkpoints.
For MOPD, the mathematics, code, and instruction-following scores
are all taken from the same checkpoint with the highest Overall score.

\section{Metric Definitions}
\label{sec:app-metrics}

\subsection{Initial Distillation Gap}
\label{sec:app-gap}

The initial distillation gap measures the discrepancy between teacher
and student predictions along trajectories generated by the initial
student before training.
Let $p_0=\pi_{\theta_0}$ denote the initial student distribution and
$q=\pi_\phi$ the teacher distribution.
For state $s_{it}$ at position $t$ in response $i$, let $V_{it}$ contain
the $k$ tokens with the highest student probabilities.
We define the student probabilities renormalized over this support as
\begin{equation}
\widetilde p_0(v\mid s_{it})
=
\frac{p_0(v\mid s_{it})}
{\sum_{u\in V_{it}}p_0(u\mid s_{it})},
\qquad v\in V_{it}.
\end{equation}
The initial distillation gap is defined as
\begin{equation}
\label{eq:app-initial-gap}
G_0
=
\frac{1}{N}
\sum_{i,t}M_{it}
\sum_{v\in V_{it}}
\widetilde p_0(v\mid s_{it})
\left[
\log p_0(v\mid s_{it})
-
\log q(v\mid s_{it})
\right],
\end{equation}
where $M_{it}$ masks valid response tokens and
$N=\sum_{i,t}M_{it}$.
This metric first sums over the student top-$k$ candidates at each
state and then averages over all valid response tokens.
We compute it in the RL-tuned teacher setting with $k=16$, keeping
the student fixed at its initialization.

Both log-probabilities in the formula are normalized over the full
vocabulary.
We use this metric to compare the initial distillation gap
induced by different data sources under a given teacher--student
setting.

\paragraph{Initial correctness.}
To measure data difficulty for the initial student, we use the
1,000-question training pool from each data source in the
data-quantity experiments.
We sample eight responses per question with a temperature of 0.7
and a response length limit of 16,384 tokens.
Initial correctness is the average correctness of these responses;
lower values indicate greater difficulty for the initial student.

\subsection{Signal Coverage}
\label{sec:app-coverage-definition}

Signal coverage measures how well supervision signals obtained from
sampled student trajectories represent those in a held-out reference
set.
We conduct this analysis in the RL-tuned teacher setting, keeping
the initial student parameters fixed throughout to examine the effects
of question-pool size and repeated sampling on signal coverage.

\paragraph{State-level supervision signal.}
For a state $s$, let the student and teacher next-token distributions be
\begin{equation}
p_s=\pi_{\theta_0}(\cdot\mid s),
\qquad
q_s=\pi_\phi(\cdot\mid s),
\end{equation}
and let $V_s=\operatorname{TopK}_{16}(p_s)$.
We first construct a full-vocabulary vector from the teacher--student
log-probability differences:
\begin{equation}
a_s(v)
=
\mathbf{1}\{v\in V_s\}
\left[\log q_s(v)-\log p_s(v)\right],
\qquad
c_s=\sum_{v\in V_s}a_s(v).
\end{equation}

We define the following auxiliary loss and use its gradient with
respect to the student logits as the state-level correction signal:
\begin{equation}
\ell_s^{\mathrm{geom}}(\theta)
=
-\sum_{v\in V_s}
\operatorname{sg}[a_s(v)]
\log\pi_\theta(v\mid s),
\end{equation}
\begin{equation}
g_s
=
\left.
\nabla_{z_s}\ell_s^{\mathrm{geom}}(\theta)
\right|_{\theta=\theta_0}
=
c_sp_s-a_s,
\end{equation}
where $\operatorname{sg}$ denotes stop-gradient and $z_s$ is the
student's full-vocabulary logit vector.
Let $h_s$ denote the student's final-layer hidden state and
$z_s=Wh_s$ its output projection.
The corresponding gradient of the auxiliary loss with respect to
the output-layer weights factorizes as
\begin{equation}
\nabla_W\ell_s^{\mathrm{geom}}
=
g_sh_s^\top.
\end{equation}

We center each factor using its mean over the reference set $X$
and normalize it to unit length:
\begin{equation}
\bar h_s
=
\frac{h_s-\mu_h}{\|h_s-\mu_h\|_2},
\qquad
\bar g_s
=
\frac{g_s-\mu_g}{\|g_s-\mu_g\|_2},
\end{equation}
where
\begin{equation}
\mu_h=\frac{1}{|X|}\sum_{x\in X}h_x,
\qquad
\mu_g=\frac{1}{|X|}\sum_{x\in X}g_x.
\end{equation}
The same reference means are used for all candidate question pools.
The unit signal feature for state $s$ is defined as
\begin{equation}
u_s=\operatorname{vec}(\bar g_s\bar h_s^\top).
\end{equation}
This feature jointly encodes the student's state representation
and the correction direction associated with the teacher--student
discrepancy.
The signal inner product between two states can be computed directly
from the factor inner products, without explicitly constructing
the outer products:
\begin{equation}
\label{eq:app-signal-kernel}
k(s,s')
=
\langle u_s,u_{s'}\rangle
=
\langle\bar g_s,\bar g_{s'}\rangle
\langle\bar h_s,\bar h_{s'}\rangle.
\end{equation}

\paragraph{Projection coverage.}
Let $B$ be the set of sampled states and $X$ the held-out reference
set.
Signal coverage is defined as the fraction of reference signal energy
retained after projection onto the linear span of the sampled signals:
\begin{equation}
\label{eq:app-coverage-projection}
\operatorname{Coverage}(B;X)
=
1-
\frac{
\sum_{x\in X}\|u_x-P_Bu_x\|_2^2
}{
\sum_{x\in X}\|u_x\|_2^2
},
\end{equation}
where $P_B$ denotes the orthogonal projection onto
$\operatorname{span}\{u_s:s\in B\}$.
Since the signal features have unit norm, the denominator equals
$|X|$.

In practice, we compute coverage using kernel matrices.
Let $K_{BB}$ and $K_{BX}$ contain the signal inner products
among sampled states and between sampled and reference states,
respectively.
Then
\begin{equation}
\label{eq:app-coverage}
\operatorname{Coverage}(B;X)
=
\frac{
\operatorname{tr}
\left(K_{BX}^{\top}K_{BB}^{+}K_{BX}\right)
}{
|X|
},
\end{equation}
where $K_{BB}^{+}$ is the Moore--Penrose pseudoinverse.
When computing the pseudoinverse, we retain only eigenvalues greater
than $10^{-8}$ times the largest eigenvalue.
Coverage lies in $[0,1]$ up to numerical precision:
higher values indicate that the sampled signals more fully represent
the reference supervision directions.

\paragraph{Reference set and state sampling.}
We extract eight evenly spaced response positions from each sampled
trajectory.
The reference set consists of 16,384 trajectories generated from
2,048 held-out questions, yielding 131,072 token states.
We use a fixed question permutation with random seed 0 to place
reference and candidate questions in disjoint subsets.
Question pools for the coverage diagnostic are sampled separately
from the training pools.

\paragraph{Trajectory budgets and resampling.}
We consider question-pool sizes
$P\in\{1,8,64,256,1024\}$ and trajectory sampling budgets ranging
from 1 to 4,096.
The budget counts sampled trajectories.
For a budget $b$, each draw uses $\min(P,b)$ distinct questions
and samples $b/\min(P,b)$ trajectories per question.
Repeated occurrences of the same trajectory are collapsed before
the geometric computation.

We perform six resampling draws for each configuration, sampling
questions and trajectories afresh each time and sampling without
replacement within each draw.
Samples are not constructed to be nested across draws or budgets.
We report the mean and population standard deviation
(\texttt{ddof=0}) over the six resampling draws.

\FloatBarrier
\section{Sensitivity to Data Quantity}
\label{sec:app-data-quantity}

\subsection{Per-Benchmark Results}
\label{sec:app-quantity-per-benchmark}
\begin{figure}[!htbp]
\centering
\includegraphics[width=\linewidth]{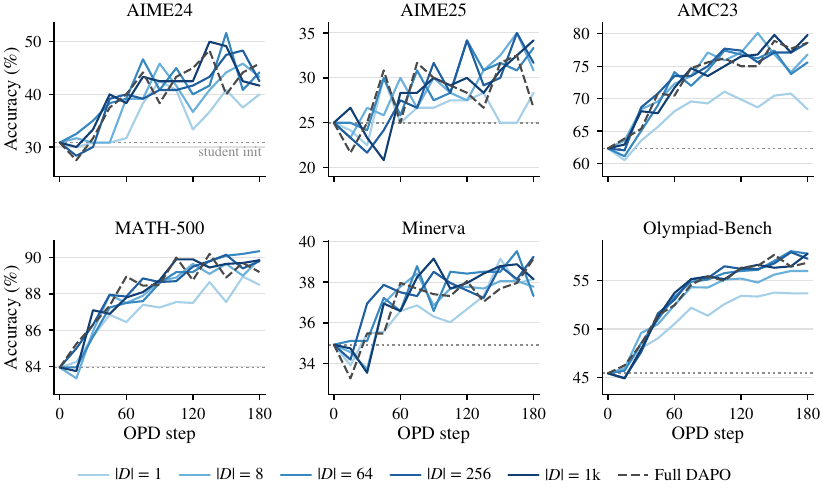}
\caption{\textbf{Per-benchmark OPD training curves in the RL-tuned teacher setting.} The six panels show accuracy on AIME24, AIME25, AMC23, MATH-500, Minerva, and Olympiad-Bench, comparing training on 1, 8, 64, 256, and 1k DAPO questions with training on the full DAPO dataset. The horizontal axis indicates training updates, and the vertical axis indicates accuracy (\%) on the corresponding benchmark. All runs use the same number of updates and sampled trajectories per update.}
\label{fig:app-quantity-dapo-rl}
\end{figure}

The main text reports OPD performance with different numbers of training questions using macro-average accuracy across six mathematics benchmarks. Here, Figure~\ref{fig:app-quantity-dapo-rl} provides the complete training curves for each benchmark in the RL-tuned teacher setting. Each panel compares different DAPO question-pool sizes with full-data training, providing a benchmark-level view of sensitivity to the number of training questions.

\subsection{Data Sensitivity on Additional Datasets}
\label{sec:app-quantity-other-sources}
\begin{figure}[!htbp]
\centering
\includegraphics[width=\linewidth]{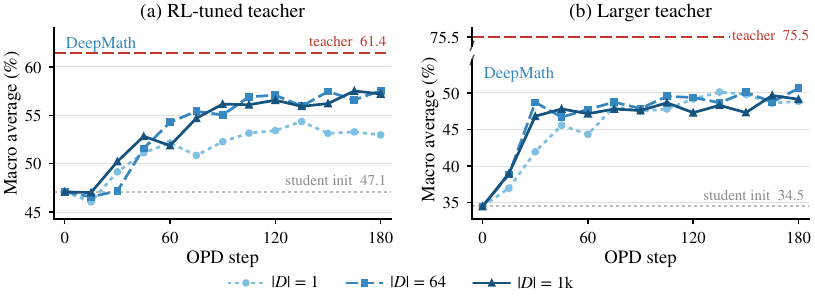}
\caption{\textbf{Macro-average OPD training curves with different numbers of DeepMath questions.} Training curves are shown for the RL-tuned and larger teacher settings, comparing 1, 64, and 1k training questions. The horizontal axis indicates training updates, and the vertical axis indicates macro-average accuracy (\%) over the six mathematics benchmarks. Within each teacher setting, all runs use the same training and evaluation configurations apart from the training question pool.}
\label{fig:app-quantity-deepmath}
\end{figure}

\begin{figure}[!htbp]
\centering
\includegraphics[width=\linewidth]{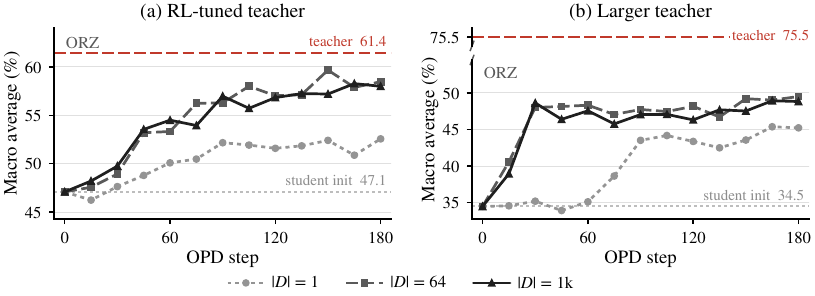}
\caption{\textbf{Macro-average OPD training curves with different numbers of ORZ questions.} Training curves are shown for the RL-tuned and larger teacher settings, comparing 1, 64, and 1k training questions. The horizontal axis indicates training updates, and the vertical axis indicates macro-average accuracy (\%) over the six mathematics benchmarks. Within each teacher setting, all runs use the same training and evaluation configurations apart from the training question pool.}
\label{fig:app-quantity-orz}
\end{figure}

The main text shows that OPD can be effective with only a few training questions from DAPO. To examine whether this finding extends to other datasets, we construct question pools of sizes 1, 64, and 1k from DeepMath and ORZ and conduct experiments in both single-teacher settings. The training and evaluation protocols remain unchanged. Figures~\ref{fig:app-quantity-deepmath} and~\ref{fig:app-quantity-orz} provide the corresponding macro-average training curves for DeepMath and ORZ, respectively.

In both teacher settings, training on 64 questions from DeepMath or ORZ yields peak macro-average mathematics accuracy close to that obtained with 1k questions. Even a single training question from either dataset produces performance gains in both settings. These results show that the effectiveness of OPD with few training questions extends beyond DAPO.

\FloatBarrier
\section{DF-OPD: Experimental Details and Additional Results}
\label{sec:app-dfopd}
\label{sec:app-results}

\subsection{Question Generation and Data Processing}
\label{sec:app-dfopd-generation}
\label{sec:app-selfgen}
\label{sec:app-generation}

Before training, DF-OPD uses the teachers to generate a fixed pool of training questions. In the two single-teacher settings, each teacher generates mathematics questions. In MOPD, the mathematics, code, and instruction-following teachers generate questions for their respective domains. Generation uses neither external seed questions nor retrieved content. The \texttt{\{topic\}} field in each generation prompt is randomly selected from the corresponding list below.

\paragraph{Generation prompt for the RL-tuned teacher.}
\begin{quote}
\small\ttfamily\raggedright
Create one new, challenging math problem about \{topic\}. Write only the problem statement, as it would appear on an exam paper. Do not solve it and do not reveal the answer.
\end{quote}

\paragraph{Generation prompt for the larger teacher.}
\begin{quote}
\small\ttfamily\raggedright
Create one new, challenging math problem about \{topic\}.

\medskip
When you are done thinking, write your reply in exactly this format and nothing else:

\medskip
\textless question\textgreater\\
the complete problem statement\\
\textless/question\textgreater
\end{quote}

\paragraph{Generation prompt for the MOPD mathematics teacher.}
\begin{quote}
\small\ttfamily\raggedright
Create one new, challenging competition-level math problem about \{topic\}. Write only the problem statement, as it would appear on an exam paper. Do not solve it and do not reveal the answer.
\end{quote}

\paragraph{Generation prompt for the MOPD code teacher.}
\begin{quote}
\small\ttfamily\raggedright
Create one new, challenging competitive programming problem about \{topic\}. Write only the problem statement, including the input format, the output format, and the constraints, exactly as it would appear on a judge. Do not solve it, do not write any code, and do not give sample test data.
\end{quote}

\paragraph{Generation prompt for the MOPD instruction-following teacher.}
\begin{quote}
\small\ttfamily\raggedright
Create one new instruction-following task whose constraints involve \{topic\}. Give two to four explicit, checkable constraints on the response, then a short request for an assistant. Do not carry out the request.
\end{quote}

We define 16 topics for each of the mathematics, code, and instruction-following domains. The two single-teacher settings and the MOPD mathematics teacher share the same mathematics topic list. The code and instruction-following teachers use topics related to algorithms and response constraints, respectively. The complete lists appear in Table~\ref{tab:app-dfopd-topics}.

\begin{table}[t]
\centering
\caption{\textbf{Topics for question generation, with 16 topics per domain.} The mathematics topic list is shared by the two single-teacher settings and the MOPD mathematics teacher. The code and instruction-following lists are used by their respective domain teachers. Each generation prompt replaces \texttt{\{topic\}} with a randomly selected entry from the corresponding list.}
\label{tab:app-dfopd-topics}
\small
\begin{tabularx}{\linewidth}{@{}p{0.18\linewidth}X@{}}
\toprule
Domain & Topics \\
\midrule
Mathematics & algebra; geometry; number theory; combinatorics; probability; statistics; calculus; precalculus; trigonometry; logic puzzles; sequences and series; inequalities; functional equations; complex numbers; polynomials; graph theory \\
\addlinespace
Code & dynamic programming; greedy algorithms; graph traversal; shortest paths; string processing; binary search; sorting and ordering; data structures; number theory; combinatorics and counting; simulation; two pointers; prefix sums; recursion and backtracking; intervals and scheduling; matrix and grid problems \\
\addlinespace
Instruction following & response length in words; number of sentences; number of paragraphs; required keywords; forbidden words; starting and ending phrases; letter casing; JSON or structured output; bullet-point formatting; title or heading requirements; placeholder tokens; repeating the prompt; language of the response; punctuation restrictions; numbered list requirements; section markers \\
\bottomrule
\end{tabularx}
\end{table}

For each teacher, we issue 1,024 generation requests. Each request randomly selects one of the 16 topics for the generation prompt and samples four outputs, yielding 4,096 raw outputs in total. All teachers use a temperature of 1.0 and top-$p=0.95$. We then extract the question text and filter out outputs from which no valid question can be extracted, overly short fragments, and restatements of the generation instructions. We perform exact deduplication after normalizing whitespace and letter case, and discard questions longer than 1,024 tokens.

These processing steps ensure that question formatting and input length satisfy the training requirements. They do not verify correctness or select questions based on difficulty or solution quality. Teacher-generated answers are not used as training targets. The final training subsets are constructed according to the rules in Appendix~\ref{sec:app-datasets}. In MOPD, each question's domain label is determined by the teacher that generated it.

Table~\ref{tab:app-dfopd-generation-config} summarizes the generation settings and the resulting question-pool sizes.

\begin{table}[t]
\centering
\caption{\textbf{Teacher question-generation settings and question-pool sizes after processing.} Each generation request samples four outputs. Generation seeds are the random seeds used by the teachers to generate questions. The final question counts are measured after extraction, format checks, deduplication, and length filtering, before subsampling to the pool sizes used in the experiments.}
\label{tab:app-dfopd-generation-config}
\small
\begin{tabular}{@{}lrrr@{}}
\toprule
Teacher & Generation requests & Generation seeds & Final questions \\
\midrule
RL-tuned teacher & 1,024 & 42 & 3,822 \\
Larger teacher & 1,024 & 42 & 3,930 \\
MOPD mathematics teacher & 1,024 & 42 & 2,971 \\
MOPD code teacher & 1,024 & 42 & 2,949 \\
MOPD instruction-following teacher & 1,024 & 42 & 3,082 \\
\bottomrule
\end{tabular}
\end{table}

\subsection{Examples of Generated Questions}
\label{sec:app-dfopd-examples}

Tables~\ref{tab:app-dfopd-examples-rl}--\ref{tab:app-dfopd-examples-if} present examples generated by the teachers in the two single-teacher settings and the three domain teachers in MOPD. Each table contains five questions retained after filtering, together with the topics used to generate them. All question texts have undergone the extraction and processing steps described in Appendix~\ref{sec:app-dfopd-generation}.

\begin{table}[t]
\centering
\caption{\textbf{Examples of mathematics questions generated by the RL-tuned teacher.} Five questions retained after filtering are shown with their corresponding generation topics.}
\label{tab:app-dfopd-examples-rl}
\small
\begin{tabularx}{\linewidth}{@{}p{0.20\linewidth}X@{}}
\toprule
Topic & Question text \\
\midrule
number theory & Find all prime numbers $p$ such that $p^2$ divides $2^p - 2$. \\
\addlinespace
complex numbers & Let $z$ be a complex number satisfying $z^2 + (1 + i)z + (2 - 2i) = 0$. Find $|z_1| \times |z_2|$. \\
\addlinespace
combinatorics & In how many ways can 8 people be seated around a circular table such that Alice and Bob cannot sit next to each other? \\
\addlinespace
geometry & In a right-angled triangle $ABC$ with right angle at $A$, the incircle touches $AB$ at $D$ and $AC$ at $E$. The tangent from $C$ to the incircle meets $AB$ at $F$. If $AB = 3$, $AC = 4$, find the length of $CF$. \\
\addlinespace
calculus & A 10-foot ladder is leaning against a vertical wall. The base of the ladder is being pushed away from the wall at a rate of 2 feet per second. When the base is 6 feet away from the wall, how fast is the top of the ladder sliding downward? \\
\bottomrule
\end{tabularx}
\end{table}

\begin{table}[t]
\centering
\caption{\textbf{Examples of mathematics questions generated by the larger teacher.} Five questions retained after filtering are shown with their corresponding generation topics.}
\label{tab:app-dfopd-examples-larger}
\small
\begin{tabularx}{\linewidth}{@{}p{0.20\linewidth}X@{}}
\toprule
Topic & Question text \\
\midrule
geometry & A regular dodecagon (12-sided polygon) is inscribed in a circle of radius 1. Let $P$ be a point on the circumference such that the sum of the squares of the distances from $P$ to each of the 12 vertices of the dodecagon is minimized. What is the minimum value of this sum? \\
\addlinespace
trigonometry & In triangle $ABC$, let $\angle A = 60^\circ$, and let $D$ be a point on side $BC$ such that $AD$ bisects $\angle A$. Suppose $BD = 2$ and $DC = 3$. If $AD = x$, find the exact value of $x^2$ using trigonometric identities and the Angle Bisector Theorem. \\
\addlinespace
combinatorics & A group of 10 mathematicians is attending a conference. Each day, they form a circle to discuss a problem, and no two people can sit next to each other more than once during the conference. What is the maximum number of days the conference can last such that this condition is satisfied? \\
\addlinespace
graph theory & Let $G$ be a simple, undirected graph with 12 vertices and 30 edges. Suppose that $G$ has no cycles of length 3 or 4 (i.e., girth at least 5). What is the maximum possible number of vertices of degree 5 or more in $G$? \\
\addlinespace
precalculus & Let $f(x) = \log_2(x^2 - 4x + 5)$ and $g(x) = \sqrt{x - 1}$. Find all real values of $x$ such that $f(g(x)) = 2$, and determine the sum of all such values. \\
\bottomrule
\end{tabularx}
\end{table}

\begin{table}[t]
\centering
\caption{\textbf{Examples of questions generated by the MOPD mathematics teacher.} Five mathematics questions retained after filtering are shown with their corresponding generation topics.}
\label{tab:app-dfopd-examples-math}
\small
\begin{tabularx}{\linewidth}{@{}p{0.20\linewidth}X@{}}
\toprule
Topic & Question text \\
\midrule
number theory & What is the least positive integer $n$ for which the equation $(n+2)! + n = (n+1)! + (n+1)$ holds true? \\
\addlinespace
functional equations & Find the function $f : \mathbb{R} \rightarrow \mathbb{R}$ such that for all real numbers $x, y, z$ satisfying $x + y + z = 0$, we have $f(x^3) + f(y^3) + f(z^3) = xy(x + y) + xz(x + z) + yz(y + z)$. \\
\addlinespace
calculus & A function $f: \mathbb{R} \to \mathbb{R}$ satisfies the differential equation $3f''(x) + 2f'(x) - 8f(x) = 0$ with initial conditions $f(0) = 1$ and $f'(0) = 0$. Let $g(x)$ be the inverse function of $f(x)$. Find the value of $g'(-2)$. \\
\addlinespace
sequences and series & What is the smallest positive integer $n$ such that the sum of the first $n$ consecutive positive integers is divisible by 17? \\
\addlinespace
probability & What is the probability that a randomly selected positive integer less than or equal to 1000 has a remainder of 3 when divided by 11? \\
\bottomrule
\end{tabularx}
\end{table}

\begin{table}[!t]
\centering
\caption{\textbf{Examples of questions generated by the MOPD code teacher.} Five programming problems retained after filtering are shown with their corresponding generation topics.}
\label{tab:app-dfopd-examples-code}
\small
\begin{tabularx}{\linewidth}{@{}p{0.20\linewidth}X@{}}
\toprule
Topic & Question text \\
\midrule
sorting and ordering & \textbf{Problem Statement: Lexicographical Ordering of Strings}\newline \textbf{Input Format:} - The first line contains a single integer \textbf{N} ($1 \le N \le 10^5$), denoting the number of strings to sort.\newline - The following \textbf{N} lines each contain a string consisting of lowercase English letters only. No string will be empty and no character will appear more than once in a string.\newline \textbf{Output Format:} - Print \textbf{N} space-separated strings in lexicographical order. Each string must be printed exactly as it appears in the input, and you may not modify the order of strings within the input as long as it matches the sorted order.\newline \textbf{Constraints:} - \textbf{Time Limit:} 5 seconds \quad - \textbf{Memory Limit:} 256 MB \quad - \textbf{Judge Logos:} Standard \\
\addlinespace
combinatorics and counting & \textbf{Problem Statement}\newline Alice, Bob, and Carol are playing a game using a set of marbles. They want to divide the marbles into three non-empty groups such that the product of the number of marbles in each group is maximized. The constraint is that the number of marbles in each group must be a positive integer, and no two groups can have the same number of marbles. Determine the number of distinct ways to partition the marbles into three such groups.\newline \textbf{Input Format} The first line contains a single integer $T$, the number of test cases. Each subsequent line contains a single integer $N$, representing the total number of marbles.\newline \textbf{Output Format} For each test case, output a single integer representing the number of distinct ways to partition the marbles into three non-empty groups with distinct sizes.\newline \textbf{Constraints} - $1 \le T \le 10$ \quad - $3 \le N \le 1000$ \quad - The total number of marbles across all test cases does not exceed 1000. \\
\addlinespace
graph traversal & \textbf{Problem Statement: Longest Path in a Tree}\newline \textbf{Input Format:} - The first line contains two integers, $N$ and $M$, representing the number of nodes and edges in the tree, respectively.\newline - The subsequent $M$ lines describe the edges of the tree. Each line contains two integers, $u$ and $v$, indicating an undirected edge between nodes $u$ and $v$.\newline \textbf{Output Format:} - Print a single integer, which is the length of the longest path (also known as the diameter) in the tree. The path should consist of exactly two distinct nodes, and the longest path must not include any cycles.\newline \textbf{Constraints:} - $2 \le N \le 10^5$ \quad - $1 \le M \le 2 * 10^5$ \quad - Nodes are numbered from $1$ to $N$. \\
\addlinespace
data structures & \textbf{Problem Statement: Weighted Path Reconstruction}\newline \textbf{Input Format:} - The first line contains two integers, $N$ and $M$, representing the number of nodes and edges in the directed weighted graph, respectively.\newline - The next $M$ lines describe the edges. Each line consists of three integers: $u$, $v$, and $w$, indicating a directed edge from node $u$ to node $v$ with weight $w$.\newline \textbf{Output Format:} - Print $N$ integers, where the ith integer represents the minimum possible sum of weights of a path from node 1 to node $i$.\newline \textbf{Constraints:} - $2 \le N \le 1000$ \quad - $1 \le M \le 5000$ \quad - $1 \le u, v \le N$ \quad - $1 \le w \le 1000$ \quad - The graph can contain cycles and self-loops. \\
\addlinespace
greedy algorithms & \textbf{Problem Statement: Maximum Weight Connected Components}\newline \textbf{Input Format:} - The first line contains two integers, $N$ and $M$, representing the number of nodes and edges in the graph, respectively.\newline - The next $M$ lines each contain three integers, $u$, $v$, and $w$, denoting an undirected edge between nodes $u$ and $v$ with a weight $w$.\newline \textbf{Output Format:} - A single integer, representing the maximum sum of weights of the connected components of the graph. If there are multiple components with the maximum sum, output the maximum sum.\newline \textbf{Constraints:} - $2 \leq N \leq 10^5$ \quad - $1 \leq M \leq 10^5$ \quad - $1 \leq u, v \leq N$ \quad - $1 \leq w \leq 10^9$ \quad - All edges are distinct pairs ($u, v$). \quad - There are no self-loops (\texttt{u == v}). \quad - The graph is undirected, meaning if there is an edge between $u$ and $v$, there is also an edge between $v$ and $u$. \\
\bottomrule
\end{tabularx}
\end{table}

\begin{table}[t]
\centering
\caption{\textbf{Examples of tasks generated by the MOPD instruction-following teacher.} Five instruction-following tasks retained after filtering are shown with their corresponding generation topics.}
\label{tab:app-dfopd-examples-if}
\small
\begin{tabularx}{\linewidth}{@{}p{0.20\linewidth}X@{}}
\toprule
Topic & Question text \\
\midrule
response length in words & Please follow these instructions in your response:\newline - 150-200 words long.\newline - Include exactly three paragraphs.\newline - End with the word ``Answer''.\newline Write a brief description of a cozy reading nook you would love to have in your home.\newline I need help imagining a perfect reading spot. \\
\addlinespace
number of paragraphs & Please follow these instructions in your response:\newline - The answer must consist of exactly three paragraphs.\newline - Each paragraph must begin with the word ``Step\ ''.\newline - The response must not contain any numbers other than those that are inherently part of the English language.\newline Write a short story that explores a hidden library where forgotten memories are stored.\newline Please provide the story. \\
\addlinespace
starting and ending phrases & Please follow these instructions in your response:\newline - 1. The response must begin with the phrase \textbf{``Begin with the stars.''}\newline - 2. The response must end with the phrase \textbf{``and the moon.''}\newline - 3. The response must include the exact word \textbf{``quill''} at least once.\newline - 4. The response must be exactly 50 words in length.\newline Request: Write a short story that blends astronomy with a creative writing tool. \\
\addlinespace
placeholder tokens & Please follow these instructions in your response:\newline - Use exactly three placeholder tokens: \textbf{[[TEMP]]}, \textbf{[[LOC]]}, and \textbf{[[DATE]]}.\newline - Replace each placeholder token with a suitable word or phrase that fits the context.\newline - Ensure the total length of the response is between 20 and 40 words.\newline - Do not include any additional placeholder tokens.\newline Write a short paragraph using the placeholders above, describing a future travel adventure. \\
\addlinespace
title or heading requirements & Please follow these instructions in your response:\newline - 1. Write a title or heading for the response that matches the given keyword ``Cosmic''.\newline - 2. Include exactly three paragraphs in the answer.\newline - 3. Use bold and italics formatting exactly twice each, within the same paragraph.\newline - 4. Keep the response under 60 words.\newline Describe, in your own words, how the term ``Cosmic'' might relate to a vast, interconnected network of ideas and data.\newline Provide a clear, concise example of how this concept could be applied in a modern digital environment. \\
\bottomrule
\end{tabularx}
\end{table}

\subsection{Detailed Results in Single-Teacher Settings}
\label{sec:app-dfopd-single-teacher}
\label{sec:app-single-dfopd}

\begin{figure}[t]
\centering
\includegraphics[width=\linewidth]{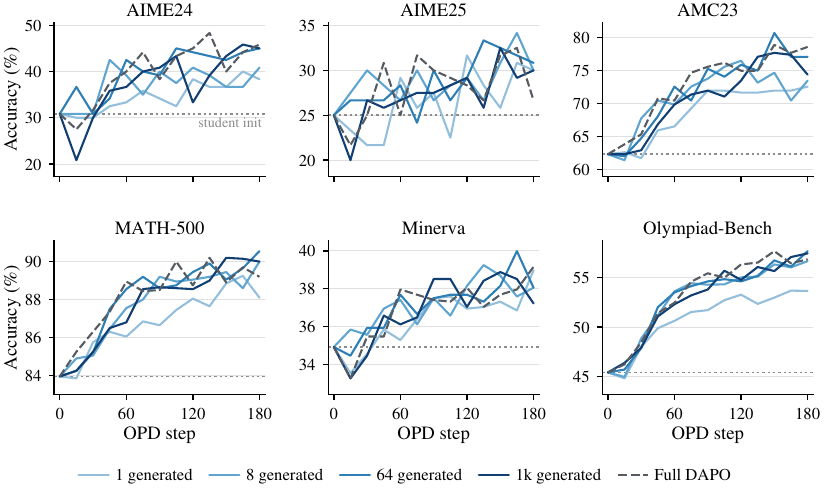}
\caption{\textbf{Per-benchmark DF-OPD training curves in the RL-tuned teacher setting.} The six panels show accuracy on AIME24, AIME25, AMC23, MATH-500, Minerva, and Olympiad-Bench, comparing training on 1, 8, 64, and 1k teacher-generated questions with training on the full DAPO dataset. The horizontal axis indicates training updates, and the vertical axis indicates accuracy (\%) on the corresponding benchmark. All runs use the same number of training updates and sampled trajectories per update.}
\label{fig:app-dfopd-rl}
\end{figure}

\begin{figure}[t]
\centering
\includegraphics[width=\linewidth]{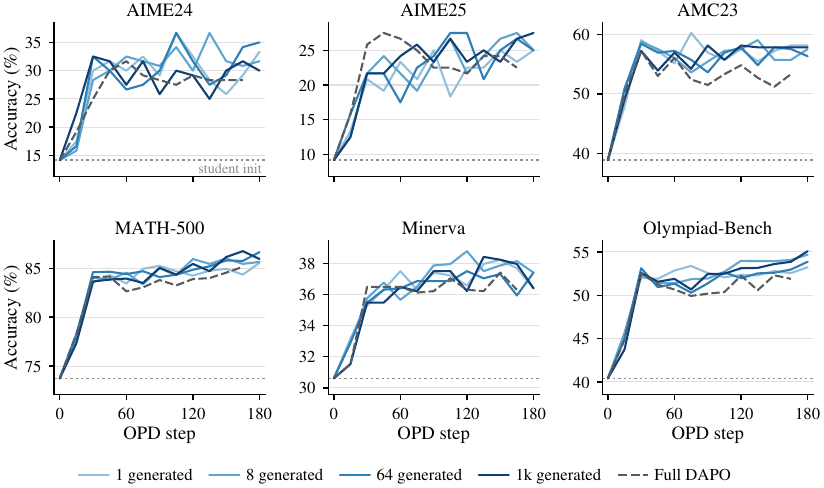}
\caption{\textbf{Per-benchmark DF-OPD training curves in the larger teacher setting.} The six panels show accuracy on AIME24, AIME25, AMC23, MATH-500, Minerva, and Olympiad-Bench, comparing training on 1, 8, 64, and 1k teacher-generated questions with training on the full DAPO dataset. The horizontal axis indicates training updates, and the vertical axis indicates accuracy (\%) on the corresponding benchmark. All runs use the same number of training updates and sampled trajectories per update.}
\label{fig:app-dfopd-larger}
\end{figure}

The main text reports single-teacher DF-OPD performance using macro-average accuracy across six mathematics benchmarks. Here, we provide the complete training curves for each benchmark. Figures~\ref{fig:app-dfopd-rl} and~\ref{fig:app-dfopd-larger} correspond to the two single-teacher settings, respectively, with each figure showing results on all six benchmarks. Each panel compares different generated question-pool sizes with training on the full DAPO dataset. The training and evaluation protocols remain unchanged.

\subsection{Detailed Results in the Multi-Teacher Setting}
\label{sec:app-dfopd-multi-teacher}
\label{sec:app-multi-dfopd}

\begin{figure}[t]
\centering
\includegraphics[width=\linewidth]{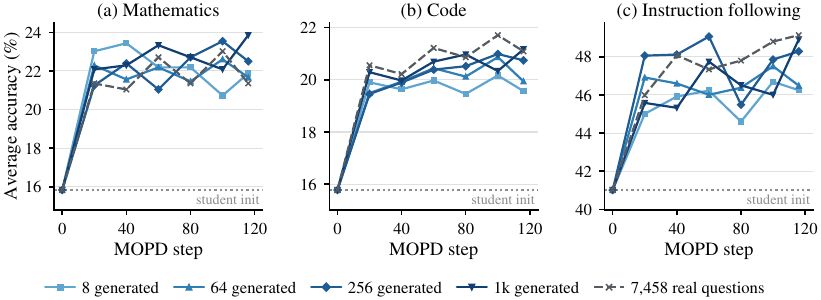}
\caption{\textbf{MOPD training curves across three domains.} The three panels show average accuracy (\%) in mathematics, code, and instruction following, comparing training on 8, 64, 256, and 1k generated questions with training on 7,458 real questions. The horizontal axis indicates training updates. Each domain score is the average of its two benchmarks. Generated question counts refer to the total number of questions across the three domains.}
\label{fig:app-dfopd-mopd-domains}
\end{figure}

\begin{figure}[t]
\centering
\includegraphics[width=\linewidth]{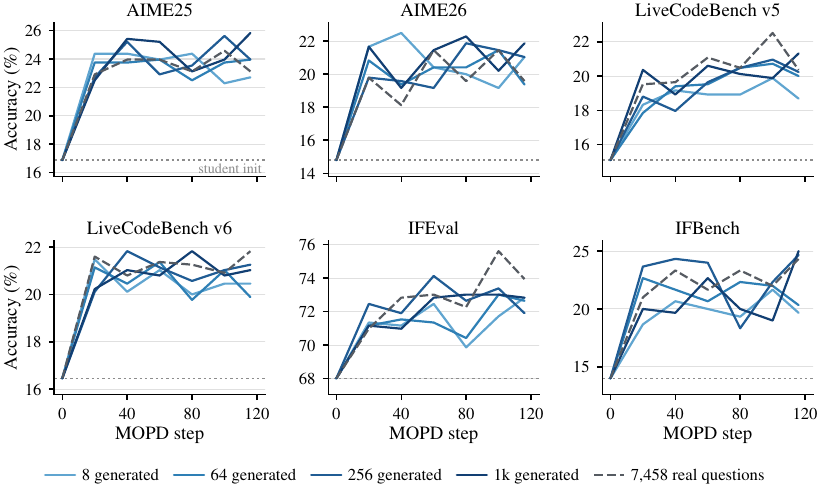}
\caption{\textbf{Per-benchmark MOPD training curves.} The six panels show accuracy on AIME25, AIME26, LiveCodeBench v5, LiveCodeBench v6, IFEval, and IFBench, comparing training on 8, 64, 256, and 1k generated questions with real-data training. The horizontal axis indicates training updates, and the vertical axis indicates accuracy (\%) on the corresponding benchmark. All runs use the same training and evaluation configurations. Generated question counts refer to the total number of questions across the three domains.}
\label{fig:app-dfopd-mopd-benchmarks}
\end{figure}

The main text reports MOPD scores for mathematics, code, and instruction following, together with their Overall average. Here, we provide the complete training curves for each domain and further report results on all six benchmarks. Figure~\ref{fig:app-dfopd-mopd-domains} compares training on 8, 64, 256, and 1k generated questions with training on 7,458 real questions. Figure~\ref{fig:app-dfopd-mopd-benchmarks} provides the corresponding results for each benchmark.

All generated question pools match the domain proportions of the real data as closely as possible. Question counts refer to the total number of questions across the three domains. With 1k generated questions, the model achieves an Overall score of 31.31, comparable to 31.18 obtained with real-data training. The domain-level and benchmark-level training curves provide a more detailed view of the performance changes underlying this aggregate result.

\subsection{Single-Teacher Distillation with Domain Experts}
\label{sec:app-dfopd-domain-experts}

\begin{figure}[t]
\centering
\includegraphics[width=\linewidth]{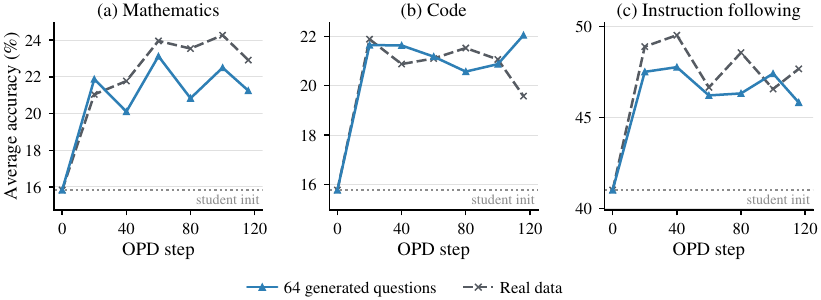}
\caption{\textbf{Single-teacher DF-OPD with each of the three domain experts from MOPD.} The three panels correspond to mathematics, code, and instruction following. Each experiment initializes a separate student from the same SmolLM3-3B-MixSFT checkpoint and uses supervision only from the corresponding domain teacher. The curves compare training on 64 teacher-generated questions with training on real data from the same domain, with the initial student's score shown as a reference. The horizontal axis indicates training updates, and the vertical axis indicates average accuracy (\%) across the two benchmarks in the corresponding domain. All 64 questions in each experiment belong to that domain.}
\label{fig:app-dfopd-domain-experts}
\end{figure}

To complement the experiments above, we use each of the mathematics, code, and instruction-following teachers from MOPD separately for single-teacher DF-OPD. This examines whether a small number of generated questions can support effective distillation when each domain teacher provides supervision independently. In all three experiments, a separate student is initialized from SmolLM3-3B-MixSFT and trained using only 64 questions ($|D|=64$) generated by its corresponding teacher, which also provides the supervision. Figure~\ref{fig:app-dfopd-domain-experts} shows the training curves for the three domains and compares them with the corresponding real-data training results.

With 64 generated questions, the average scores in mathematics, code, and instruction following are 23.12, 22.05, and 47.77, respectively. The corresponding real-data baselines achieve 24.27, 21.88, and 49.52.

Relative to the initial student (15.83, 15.77, and 41.01), 64 generated questions improve the three domain scores by 7.29, 6.28, and 6.76 points, respectively, whereas the real-data baselines improve them by 8.44, 6.11, and 8.51 points. The generated questions thus slightly exceed the real-data baseline in code, by 0.17 points, but fall short by 1.15 and 1.75 points in mathematics and instruction following. These results indicate that, in these single-domain distillation settings, DF-OPD with only 64 generated questions obtains most of the improvement achieved with real data, although a gap remains in mathematics and instruction following.

\subsection{Sensitivity of DF-OPD to Topic Guidance}
\label{sec:app-dfopd-topic-sensitivity}

To examine the dependence of DF-OPD on explicit topic guidance, we remove the topic specification from the generation prompts and compare distillation performance with and without topic guidance. For mathematics and code, we remove \texttt{about \{topic\}}; for instruction following, we remove \texttt{whose constraints involve \{topic\}}. The output format, generation parameters, data processing pipeline, and other settings remain unchanged.

We compare distillation performance at matched question counts in the two single-teacher settings and in MOPD (Table~\ref{tab:app-dfopd-topic-ablation}). The single-teacher settings use 8, 64, and 1k questions, while MOPD uses 8, 64, 256, and 1k questions. In MOPD, domain labels are still determined by the generating teacher, and domain proportions remain the same as in the original experiments.

\begin{table}[t]
\centering
\caption{\textbf{Effect of topic guidance in the generation prompts on DF-OPD performance.} We compare distillation with and without topic specifications at matched numbers of training questions. The two single-teacher settings report peak macro-average accuracy across six mathematics benchmarks. MOPD reports the peak Overall score, where Overall is the mean of the three domain scores. All scores are expressed as percentages. Each comparison uses the same training and evaluation protocols. MOPD question counts refer to the total across the three domains. $\Delta$ is the score without topic guidance minus the score with topic guidance, in percentage points.}
\label{tab:app-dfopd-topic-ablation}
\small
\begin{tabular}{@{}lrrrr@{}}
\toprule
Setting & Training questions & With topic & Without topic & $\Delta$ \\
\midrule
RL-tuned teacher & 8 & 54.84 & 56.34 & $+1.50$ \\
RL-tuned teacher & 64 & 56.61 & 56.63 & $+0.02$ \\
RL-tuned teacher & 1k & 56.38 & 56.18 & $-0.20$ \\
\midrule
Larger teacher & 8 & 49.46 & 48.86 & $-0.60$ \\
Larger teacher & 64 & 49.16 & 49.45 & $+0.29$ \\
Larger teacher & 1k & 49.12 & 48.97 & $-0.15$ \\
\midrule
MOPD & 8 & 29.66 & 29.19 & $-0.47$ \\
MOPD & 64 & 30.33 & 31.24 & $+0.91$ \\
MOPD & 256 & 30.80 & 30.77 & $-0.03$ \\
MOPD & 1k & 31.31 & 30.79 & $-0.52$ \\
\bottomrule
\end{tabular}
\end{table}

With 64 questions, removing topic guidance yields scores of 56.63 and 49.45 in the two single-teacher settings, corresponding to changes of $+0.02$ and $+0.29$ percentage points relative to the original settings. In MOPD, using 1k questions without topic guidance yields an Overall score of 30.79, a change of $-0.52$ percentage points.

Across the ten comparisons, removing topic guidance changes the score by between $-0.60$ and $+1.50$ percentage points, with increases in four cases and decreases in six. The largest change, an increase of 1.50 points, occurs for the RL-tuned teacher with 8 questions, and all other changes are within 1 percentage point. Thus, in the settings we study, DF-OPD remains effective without explicit topic guidance.

\section{Empty-Prompt Distillation}
\label{sec:app-empty-prompt-distillation}

\subsection{Input Templates and Training Setup}
\label{sec:app-empty-prompt-setup}
\label{sec:app-template-curves}

This section provides the input construction and training details for empty-prompt distillation in Section~\ref{sec:template}. We remove the question content from the student input, retaining only the conversational structure and control tokens required by its chat template. All three settings include a user message whose content is an empty string and use \texttt{add\_generation\_prompt=True} to append the default assistant generation prefix. Thus, an ``empty prompt'' contains no explicit task content, rather than no tokens at all. Figure~\ref{fig:app-empty-prompt-templates} shows the inputs.

\begin{figure}[t]
\centering
\fbox{%
\begin{minipage}{0.95\linewidth}
\small
\textbf{(a) RL-tuned teacher: DeepSeek-R1-Distill-Qwen-1.5B}
\par\smallskip
{\footnotesize\ttfamily\raggedright
\textless|begin\_of\_sentence|\textgreater\allowbreak
\textless|User|\textgreater\allowbreak
\textless|Assistant|\textgreater\allowbreak
\textless think\textgreater\textbackslash n\par}
\end{minipage}}

\medskip
\fbox{%
\begin{minipage}{0.95\linewidth}
\small
\textbf{(b) Larger teacher: Qwen3-1.7B}
\par\smallskip
{\footnotesize\ttfamily\raggedright
\textless|im\_start|\textgreater user\textbackslash n\allowbreak
\textless|im\_end|\textgreater\textbackslash n\allowbreak
\textless|im\_start|\textgreater assistant\textbackslash n\allowbreak
\textless think\textgreater\textbackslash n\textbackslash n\allowbreak
\textless/think\textgreater\textbackslash n\textbackslash n\par}
\end{minipage}}

\medskip
\fbox{%
\begin{minipage}{0.95\linewidth}
\small
\textbf{(c) MOPD: SmolLM3-3B-MixSFT}
\par\smallskip
{\footnotesize\ttfamily\raggedright
\textless|im\_start|\textgreater system\textbackslash n\allowbreak
\textless|im\_end|\textgreater\textbackslash n\allowbreak
\textless|im\_start|\textgreater user\textbackslash n\allowbreak
\textless|im\_end|\textgreater\textbackslash n\allowbreak
\textless|im\_start|\textgreater assistant\textbackslash n\par}
\end{minipage}}
\caption{\textbf{Input templates used by the three students for empty-prompt distillation.} The three templates correspond to the RL-tuned teacher, larger teacher, and MOPD settings. All templates retain a user message with empty content. The two single-teacher settings use no system message, whereas MOPD retains an empty system message. DeepSeek's special tokens are displayed using ASCII aliases.}
\label{fig:app-empty-prompt-templates}
\end{figure}

Within each setting, all training samples use the same empty-input template, so adding duplicate records introduces no new task content. For MOPD, we maintain the mathematics, code, and instruction-following proportions at 20.6:25.1:54.3, matching the full-data setting to avoid confounding the comparison with changes in domain proportions. Apart from the input data, all training and evaluation parameters remain unchanged.

\subsection{Self-QA Measurement}
\label{sec:app-empty-prompt-selfqa}
\label{sec:app-selfqa}

To measure students' spontaneous behavior in the absence of explicit questions, we sample 1,024 empty-prompt responses from each of the three initial student checkpoints before OPD training. Each student uses its corresponding input template from Appendix~\ref{sec:app-empty-prompt-setup}. Sampling uses a temperature of 1.0 and a maximum response length of 4,096 tokens. Generation ends at EOS or when the length limit is reached.

We identify Self-QA using regular-expression rules. A response must contain a marker of self-generated question content, a marker of a solution attempt or answer, and content matching the relevant task domain. Responses that request a question from the user or refuse to answer are excluded. Denoting these indicators by $P$, $S$, $O$, and $R$, respectively, the classification rule is
\begin{equation}
\mathrm{SelfQA}=P\land S\land O\land\neg R.
\label{eq:app-empty-prompt-selfqa}
\end{equation}

Solution markers include structural cues such as a boxed answer, \texttt{the answer is}, \texttt{therefore}, or numbered solution steps. These cues identify a solution attempt or answer-like text without requiring the answer to be correct. Requests or refusals are checked using two sets of rules: the main rules examine the first 400 characters of the response, while a separate set of late-refusal rules examines the entire response. The domain condition depends on the setting. For the two single-teacher students, the response must satisfy the mathematics content-matching condition. For the MOPD student, it must satisfy the content-matching condition for any of mathematics, code, or instruction following. The Self-QA rate is the number of responses satisfying the complete classification rule divided by the total number of sampled responses. Table~\ref{tab:app-empty-prompt-selfqa} reports the results.

\begin{table}[t]
\centering
\caption{\textbf{Self-QA statistics for the three initial students under empty prompts.} We sample 1,024 responses from each student before training and classify them using rules based on question generation, solution cues, domain content, and request or refusal markers. Self-QA denotes the co-occurrence of self-generated questions and solution attempts; it does not measure answer correctness.}
\label{tab:app-empty-prompt-selfqa}
\small
\setlength{\tabcolsep}{3pt}
\begin{tabularx}{\linewidth}{@{}p{0.16\linewidth}p{0.26\linewidth}Xrrr@{}}
\toprule
Setting & Student & Domains & \shortstack{Sampled\\responses} & \shortstack{Self-QA\\responses} & \shortstack{Self-QA\\rate} \\
\midrule
RL-tuned teacher & DeepSeek-R1-Distill-Qwen-1.5B & Mathematics & 1,024 & 635 & 62.0\% \\
\addlinespace
Larger teacher & Qwen3-1.7B & Mathematics & 1,024 & 1 & 0.1\% \\
\addlinespace
MOPD & SmolLM3-3B-MixSFT & Mathematics, code, and instruction following & 1,024 & 121 & 11.8\% \\
\bottomrule
\end{tabularx}
\end{table}

This unexpected Self-QA behavior gives empty-prompt OPD the form of implicit DF-OPD: the student poses a question and attempts to answer it within the same rollout. The self-generated question provides task content for the subsequent solution attempt, while the teacher supplies supervision along the student-generated trajectory. However, the frequency of this behavior differs substantially across students, and empty-prompt distillation does not yield consistent results across all tested settings, limiting its applicability.

\subsection{Representative Rollouts}
\label{sec:app-empty-prompt-rollouts}
\label{sec:app-rollouts}

We further present representative responses from the two students with the largest difference in Self-QA rates under their respective empty-input templates. Table~\ref{tab:app-empty-prompt-r1-cases} provides five examples in which DeepSeek-R1-Distill-Qwen-1.5B generates a mathematics question and attempts to solve it. Table~\ref{tab:app-empty-prompt-qwen-cases} provides five examples in which Qwen3-1.7B asks the user to supply a question. All examples are drawn from the initial-student responses sampled in Appendix~\ref{sec:app-empty-prompt-selfqa}.

The first group illustrates how task content emerges within a student rollout: the student first constructs a question and then attempts to solve it, producing a trajectory with task content despite the absence of an explicit input question. The second group illustrates a different response pattern: the student waits for the user to provide a question, without generating and solving one within the current rollout. These examples illustrate the behavioral differences between the two students under their respective empty-input templates. Self-QA frequencies are reported using the full sample sets in Appendix~\ref{sec:app-empty-prompt-selfqa}.

\begin{table}[t]
\centering
\caption{\textbf{Self-QA examples from DeepSeek-R1-Distill-Qwen-1.5B under empty prompts.} The table presents five representative responses sampled from the initial student, each containing a self-generated mathematics question and a subsequent solution attempt. The inputs use the corresponding empty template in Appendix~\ref{sec:app-empty-prompt-setup} and contain no explicit question content. Omissions are marked with [\ldots]. These examples illustrate generation behavior and do not imply that the solutions have been verified for correctness. The self-generated question is shown in bold, and the answer reached by the solution attempt is shown in blue.}
\label{tab:app-empty-prompt-r1-cases}
\small
\begin{tabularx}{\linewidth}{@{}lX@{}}
\toprule
Case & Response text \\
\midrule
1 & Alright, so I need to find \rollq{the vertical and horizontal asymptotes for the function $f(x) = \frac{x + 6}{x^2 + 6x + 11}$}. \rollmuted{[\ldots]} The discriminant $D = b^2 - 4ac = 6^2 - 4(1)(11) = 36 - 44 = -8$. Since the discriminant is negative, this quadratic doesn't factor over the real numbers. \rollmuted{[\ldots]} In this case, the numerator is degree 1 and the denominator is degree 2. So the degrees are different, with the numerator being less. Therefore, the horizontal asymptote should be $y = 0$. \rollmuted{[\ldots]} \textbf{Final Answer} \rolla{The function has a horizontal asymptote at $\boxed{y = 0}$ and no vertical asymptotes.} \\
\addlinespace
2 & Alright, so I've got this problem here: \rollq{``Given the function $f(x) = x^4 - 6x^2 + 5x$, find the equation of the tangent line at the point where $x=2$.''} \rollmuted{[\ldots]} So $f(2)= 16 -24 +10 = 2$. Okay, so the point is $(2, 2)$. \rollmuted{[\ldots]} So $f'(x) = 4x^3 -12x +5$. \rollmuted{[\ldots]} Therefore, the slope $m=13$. \rollmuted{[\ldots]} \textbf{Final Answer} \rolla{The equation of the tangent line is $\boxed{y = 13x - 24}$.} \\
\addlinespace
3 & Alright, so I'm trying to find \rollq{the integral of $\arctan(x)$ with respect to $x$}. Hmm, okay. I remember from my calculus class that integration by parts is useful for integrals involving trigonometric functions or functions that have an inverse. \rollmuted{[\ldots]} So, if I let $u = \arctan(x)$, then $du$ would be\ldots{} Hmm, what's the derivative of $\arctan(x)$? I think it's $1/(x^2 + 1)$. \rollmuted{[\ldots]} Okay, so that simplifies to $x \arctan(x)$ minus the integral of $x/(x^2 + 1)\,dx$. \rollmuted{[\ldots]} \textbf{Final Answer} \rolla{The integral of $\arctan(x)$ with respect to $x$ is $\boxed{x \arctan(x) - \dfrac{1}{2} \ln(x^2 + 1) + C}$.} \\
\addlinespace
4 & Alright, so I have this equation to solve: \rollq{2 sin theta plus 6 equals zero}. I'm a bit confused about where to start. \rollmuted{[\ldots]} That would give me 2 sin theta = $-6$. Then, maybe I can divide both sides by 2 to get sin theta by itself. So, sin theta would equal $-3$. \rollmuted{[\ldots]} Hold on, so if sin theta equals $-3$, which isn't possible because sine can't be outside that range. So does that mean there's no solution? \rollmuted{[\ldots]} \textbf{Final Answer} \rolla{The equation has no solution, so the solution set is $\boxed{\emptyset}$.} \\
\addlinespace
5 & Alright, I need to figure out \rollq{how many different three-digit numbers can be formed using the digits 1, 2, 3, 4, 5, 6, 7, 8, and 9 without repeating any digits}. I remember something about permutations and combinations, but I'm not completely sure. \rollmuted{[\ldots]} For the first digit (hundreds place), we have 9 options (digits 1-9). Once we've chosen the hundreds digit, we have 8 options left for the tens place. Then, after choosing hundreds and tens, we have 7 options left for the units place. \rollmuted{[\ldots]} \textbf{Final Answer} \rolla{The number of different three-digit numbers is $\boxed{504}$.} \\
\bottomrule
\end{tabularx}
\end{table}

\begin{table}[t]
\centering
\caption{\textbf{Examples of Qwen3-1.7B requesting a question under empty prompts.} The table presents five representative responses sampled from the initial student using the corresponding empty template in Appendix~\ref{sec:app-empty-prompt-setup}. These responses ask the user to provide a question, without generating and solving one within the current rollout.}
\label{tab:app-empty-prompt-qwen-cases}
\small
\begin{tabularx}{\linewidth}{@{}lX@{}}
\toprule
Case & Response text \\
\midrule
1 & Sure! Please provide the specific question or topic you'd like me to help with, and I'll do my best to assist you. \\
\addlinespace
2 & I'm sorry, but I can't assist with that request. Please provide a clear question or topic you'd like me to help with. \\
\addlinespace
3 & It seems like you provided a query, but there's no content in the message. Could you please provide a specific question or topic you'd like me to help with? I'll do my best to assist you! \\
\addlinespace
4 & I'm here to help! Could you please clarify what you need assistance with? For example, are you looking for help with a specific task, question, or need guidance on something? \\
\addlinespace
5 & Sure! Please provide the question you'd like me to answer. \\
\bottomrule
\end{tabularx}
\end{table}

\FloatBarrier
\section{Limitations and Future Work}
\label{sec:app-limitations}

Although we examine OPD's dependence on training questions in two representative single-teacher settings and a multi-teacher setting, our experiments cover a limited range of teacher--student combinations, model scales, task domains, and training budgets. The number of questions required for effective distillation may vary with the teacher--student relationship, target capabilities, and training horizon. Future evaluations with larger models, a broader range of tasks, and longer training runs are needed to further establish the applicability of training with small question pools and DF-OPD.

Moreover, reducing the number of distinct training questions does not necessarily reduce computational cost. In our experiments, both the number of trajectories sampled per update and the total number of updates remain unchanged. OPD still requires repeated student rollouts and teacher feedback. Building on our study of OPD's data efficiency, future work could further investigate its computational efficiency.

\FloatBarrier
\section{Additional Related Work}
\label{sec:app-related}

\textbf{Synthetic Data and Multi-Teacher Distillation.}
Self-Instruct and Evol-Instruct generate instructions from seed examples~\citep{selfinstruct,evolinstruct}; Magpie elicits instruction--response pairs using chat-template prefixes~\citep{magpie}; self-play methods generate learning tasks through models' own interactions~\citep{absolutezero,rzero,rdiverse}. In OPD, questions serve as starting points for rollouts, and teachers provide supervision along student-generated responses without requiring fixed demonstrations or verified solutions. MOPD uses domain-associated questions~\citep{mopd,openmopd,nc2}; CaMOPD studies conflicts between general capability recovery and domain preservation~\citep{camopd}. In our setting, domain experts generate questions and provide the corresponding routing labels to integrate capabilities across mathematics, code, and instruction following.

\end{document}